\documentclass{article}
\pdfoutput=1
\usepackage[preprint]{neurips_2023}

\usepackage[utf8]{inputenc} %
\usepackage[T1]{fontenc}    %
\usepackage{xurl}
\usepackage{hyperref}            %
\usepackage{breakurl}          %
\usepackage{amsfonts}       %
\usepackage{amsmath}       %
\usepackage{nicefrac}       %
\usepackage{microtype}      %
\usepackage{lipsum}
\usepackage{graphicx}
\usepackage{wrapfig}
\usepackage[size=small]{caption}
\usepackage{mathtools}
\usepackage{amssymb}
\usepackage{amsthm}
\usepackage[algoruled,boxed,lined,noend]{algorithm2e}
\usepackage{adjustbox} 
\usepackage{float}
\usepackage{amssymb}
\usepackage{pifont}
\usepackage{subcaption}
\usepackage{fontawesome}
\usepackage{multirow}
\usepackage{rotating}
\usepackage{array}    

\usepackage[numbers]{natbib}

\usepackage[dvipsnames]{xcolor}
\usepackage{booktabs}
\usepackage{multirow}
\usepackage{tabularx}

\definecolor{cvprblue}{rgb}{0.21,0.49,0.74}
\definecolor{citecolor}{HTML}{0071BC}
\definecolor{linkcolor}{HTML}{ED1C24}
\usepackage[capitalize]{cleveref}
\crefname{section}{Sec.}{Secs.}
\crefname{table}{Table}{Tables}
\crefname{figure}{Fig.}{Figs.}
\usepackage[most]{tcolorbox}
\usepackage{float}
\usepackage{xspace}
\tcbset{
  aibox/.style={
    width=394.18663pt,
    top=10pt,
    colback=white,
    colframe=black,
    colbacktitle=black,
    enhanced,
    center,
    attach boxed title to top left={yshift=-0.1in,xshift=0.15in},
    boxed title style={boxrule=0pt,colframe=white,},
  }
}
\newtcolorbox{AIbox}[2][]{aibox,title=#2,#1}

\usepackage{colortbl}
\definecolor{qual-fig-green}{RGB}{0,144,11}
\definecolor{qual-fig-red}{RGB}{238,0,0}
\definecolor{qual-fig-purple}{RGB}{153,51,255}

\title{Triad: Vision Foundation Model for 3D Magnetic Resonance Imaging}

\author{
Shansong Wang$^{1}$  \quad Mojtaba Safari$^{1}$ \quad Qiang Li$^{1}$ \quad Chih-Wei Chang$^{1}$ \quad Richard LJ Qiu$^{1}$ \\ \quad \textbf{Justin Roper}$^{1}$\quad \textbf{David S. Yu}$^{1}$\quad \textbf{Xiaofeng Yang}$\textsuperscript{1, \faEnvelope}$ 
\vspace{3mm}
\\
$^1$Department of Radiation Oncology, Winship Cancer Institute, Emory University School of Medicine\\
\faEnvelope \quad Corresponding author: xiaofeng.yang@emory.edu \\
}

\begin{document}
\maketitle
\begin{abstract}
	Vision foundation models (VFMs) are pre-trained on extensive image datasets to learn general representations for diverse types of data. These models can subsequently be fine-tuned for specific downstream tasks, significantly boosting performance across a broad range of applications. However, existing vision foundation models that claim to be applicable to various clinical tasks are mostly pre-trained on 3D computed tomography (CT), which benefits from the availability of extensive 3D CT databases. Significant differences between CT and magnetic resonance imaging (MRI) in imaging principles, signal characteristics, and data distribution may hinder their practical performance and versatility in MRI-specific applications. Here, we propose \textbf{Triad}, a vision foundation model for 3D MRI. Triad adopts a widely used autoencoder architecture to learn robust representations from 131,170 3D MRI volumes and uses organ-independent imaging descriptions to constrain the semantic distribution of the visual modality. The above pre-training dataset is called Triad-131K, which is currently the largest 3D MRI pre-training dataset. We evaluate Triad across three tasks, namely, organ/tumor segmentation, organ/cancer classification, and medical image registration, in two data modalities (within-domain and out-of-domain) settings using 25 downstream datasets. By initializing models with Triad's pre-trained weights, nnUNet-Triad improves segmentation performance by 2.51\% compared to nnUNet-Scratch across 17 datasets. Swin-B-Triad achieves a 3.97\% improvement over Swin-B-Scratch in classification tasks across five datasets. SwinUNETR-Triad improves by 4.00\% compared to SwinUNETR-Scratch in registration tasks across two datasets. Our study demonstrates that pre-training can improve performance when the data modalities and organs of upstream and downstream tasks are consistent. This work highlights the value of large-scale pre-training techniques for downstream tasks in 3D MRI. By open-sourcing Triad's weights, code, and data, we aim to enhance the adaptability and reliability of AI solutions for 3D MRI in clinical tasks.

\end{abstract}
\section{Introduction}
\label{sec:intro}
Each year, over 40 million magnetic resonance imaging (MRI) scans are performed in the United States, an average of 107.5 scans per 1,000 persons~\cite{oecd_mri_units_2023,shah2023review}. Globally, the annual total of MRI scans ranges between 100 to 150 million~\cite{mri_number_worldwide}. 
This has led to a growing demand for automated analysis tools~\cite{https://doi.org/10.1002/mp.16844}. In recent years, Foundation Model (FM)-driven image analysis has shown significant advancements. However, these foundation models have primarily been tailored for general computer vision tasks and are trained on numerous natural image datasets to acquire general representations applicable to a wide range of data~\cite{Caron_2021_ICCV,he2022masked,oquab2023dinov2}. They can then be fine-tuned for various specific downstream tasks, leading to significant enhancements in performance across different applications. This paradigm shift has also been widely adopted in clinical modalities (including 2D and 3D data), which has demonstrated notable improvements~\cite{moor2023foundation,chen2024towards,Zhao2024,Safari_2024}.

\begin{figure}[t] 
	\centering
	\includegraphics[width=\textwidth]{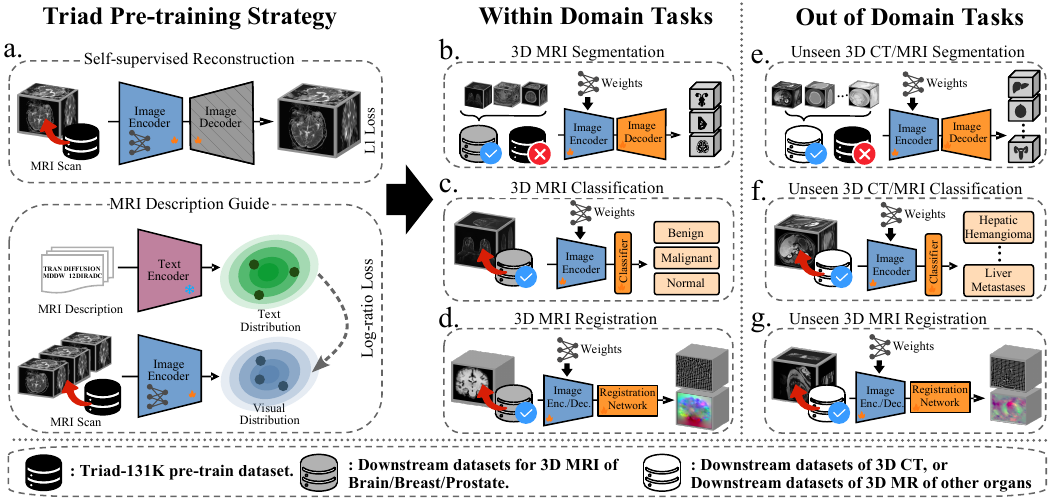}
	\caption{\textbf{Overview of Triad training and evaluation.} a. Triad pre-training strategy. Triad implements the reconstruction task based on autoencoders and uses L1 loss for optimization. Imaging descriptions are embedded into vector space to form a distribution, which serves as a supervisory signal to constrain the distribution of visual modalities using Log-ratio loss~\cite{kim2019deep}. The two losses are optimized simultaneously in a multi-task manner. Triad is then evaluated across within-domain tasks and out-of-domain tasks. These include within-domain 3D MRI segmentation, classification, and registration tasks (tasks b, c, and d). And unseen 3D CT/MRI segmentation, classification, and registration tasks (tasks e, f, and g).}
	\label{fig:overview} 
\end{figure}
\begin{figure}[!h] 
	\centering
	\includegraphics[width=\textwidth]{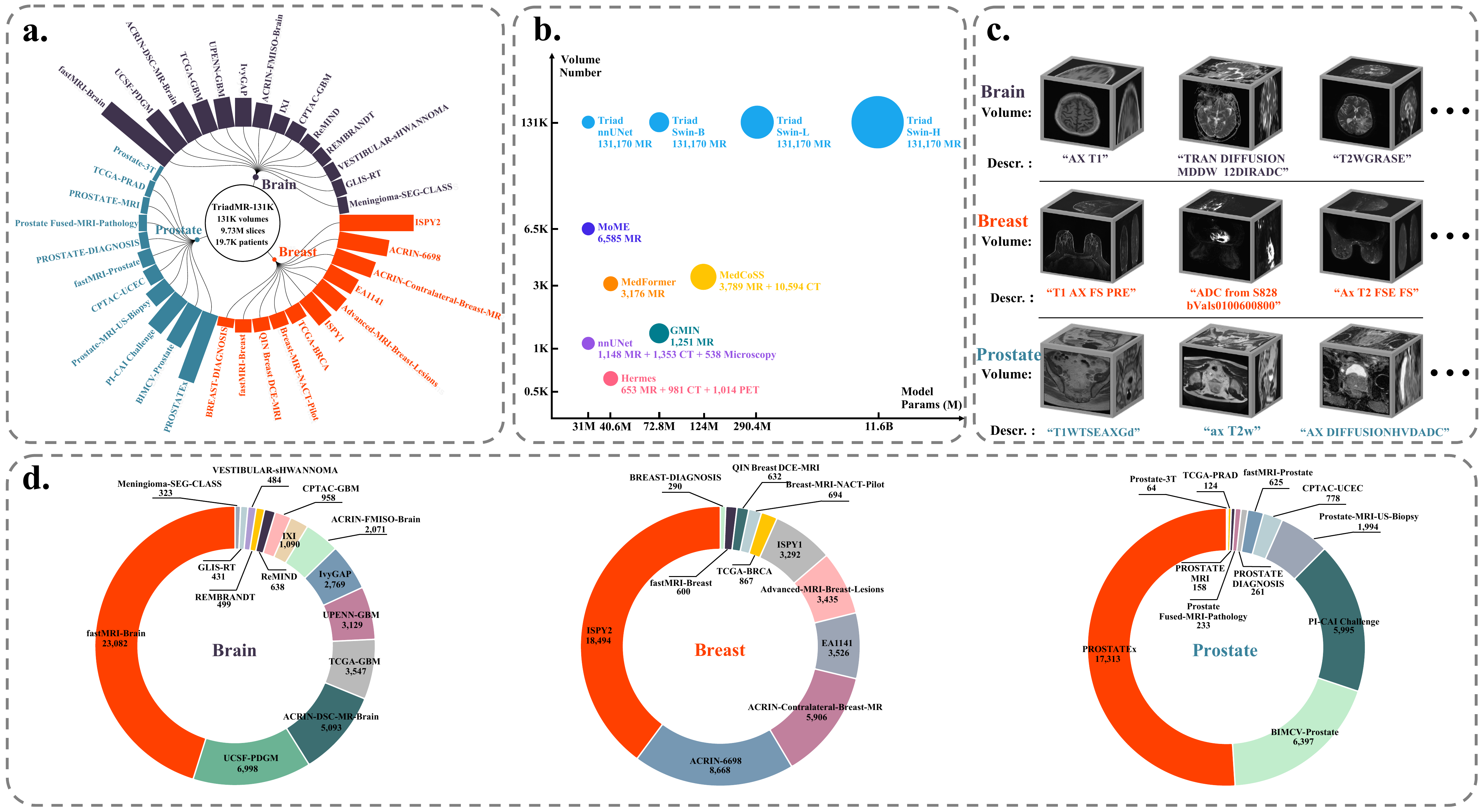}
	\caption{\textbf{An overview of the Triad-131K pre-training dataset.} a. Describes the name and scale distribution of each dataset in Triad-131K. b. We compare the parameter scale and data scale used by Triad and existing foundation models, and it is obvious that Triad  surpasses the existing models on both scales. c. Shows examples of visual volumetric modality and textual modality in Triad-131K. d. Shows the dataset scale distribution of three organs: brain, breast, and prostate.}
	\label{fig:maindataset} 
\end{figure}

Currently, the potential of foundational models specifically for 3D MRI remains largely unexplored. There are two key limitations:
Firstly, although previous general medical foundation models assert the capability to generalize to 3D MRI, substantial differences in imaging principles, signal characteristics, and data distribution between other modalities and MRI may hinder their practical performance and generalizability in MRI-specific applications~\cite{perez2021torchio,cardoso2022monai,https://doi.org/10.1002/mp.17675}. For instance, notable general medical foundation models~\cite{wu2024voco,li2024well,tang2022self} have predominantly been pre-trained on 3D computed tomography (CT). Although some foundation models leverage mixed modalities during the pre-training phase, including MRI, CT, positron emission tomography, and microscopy~\cite{ye2024continual,gao2024training,isensee2021nnu}, there is extreme data imbalance across the imaging modalities. For example, MedCoSS~\cite{ye2024continual} relied on only 3,000 MRI scans, compared to 10,000 CT scans.
Secondly, while there have been attempts to develop 3D MRI-specific foundation models, these efforts have typically focused on data from a single organ~\cite{cox2024brainsegfounder, kim2023empirical}, primarily emphasizing on T1- (T1-w) and T2-weighted (T2-w) images while overlooking the additional information of other MRI sequences. For instance, Brainsegfounder~\cite{cox2024brainsegfounder} uses brain T1-w and T2-w images from approximately 80,000 healthy subjects for pre-training, followed by self-supervised fine-tuning on specific downstream datasets. Moreover, text-based reports~\cite{blankemeier2024merlin,zhang2024generalist}, which are commonly employed as auxiliary information in 3D CT pre-trained models, are often lacking for 3D MRI data, further limiting the development of comprehensive models.
Therefore, the main challenge in building a general 3D MRI vision foundation model is collecting and curating a sufficiently large and diverse dataset. This dataset must cover a wide range of imaging modalities and hardware specifications to ensure the robustness and generalizability of the model.

To address the above limitations, we introduce Triad${^\ddagger}$\footnote[0]{${\ddagger}$ In philosophy, a Triad refers to three closely related and inseparable elements. Here, it means that the vision foundation model trained on data from three organs can robustly generalize to downstream tasks involving other modalities and organs.}, a training strategy and general vision foundation model for 3D MRI. Triad is  trained on a large-scale dataset of 131,170 3D MRI derived from 19,721 patients across 36 clinical datasets. This comprehensive dataset, termed TriadMR-131K, encompasses a wide range of 3D MRI data from three organs: breast, brain, and prostate. It includes various imaging modalities such as  T1-w, T2-w, fluid-attenuated inversion recovery (FLAIR), diffusion-weighted imaging (DWI), functional MRI (fMRI), dynamic contrast-enhanced MRI (DCE-MRI).
As illustrated in Fig. \ref{fig:maindataset} (c), we have assigned an imaging description to each MRI volume, detailing the imaging modality and associated device parameters, which adds semantics across organs.
During pre-training, we adopt a widely used autoencoder architecture to learn robust representations from this extensive and diverse dataset. In addition, we leverage organ-independent imaging descriptions to constrain the semantic distribution of visual modalities. We pre-train encoders with varying parameter sizes (ranging from 31M to 11.8B) to accommodate downstream tasks of different scales.
During fine-tuning, we demonstrate our model's versatility by replacing the decoder with task-specific adapters tailored for various 3D MRI applications. These adapters include linear classifiers for disease diagnosis, convolutional decoders for organ/tumor segmentation, and upsampling decoders for recovering deformation fields in registration tasks.
Furthermore, we extend Triad to downstream tasks involving unseen 3D CT and MRI, which we refer to as out-of-domain tasks. Our findings indicate that by combining Triad with different adapters, we not only achieve state-of-the-art performance on various within-domain tasks but also significantly outperform baselines on multiple out-of-domain tasks.
These results highlight Triad's adaptability to downstream tasks, demonstrating its potential as a versatile and efficient tool for diverse clinical applications. This adaptability could pave the way for improving the performance of various clinical tasks applied to 3D MRI.

\section{Results}

\begin{figure}[!ht] 
	\centering
	\includegraphics[width=0.95\textwidth]{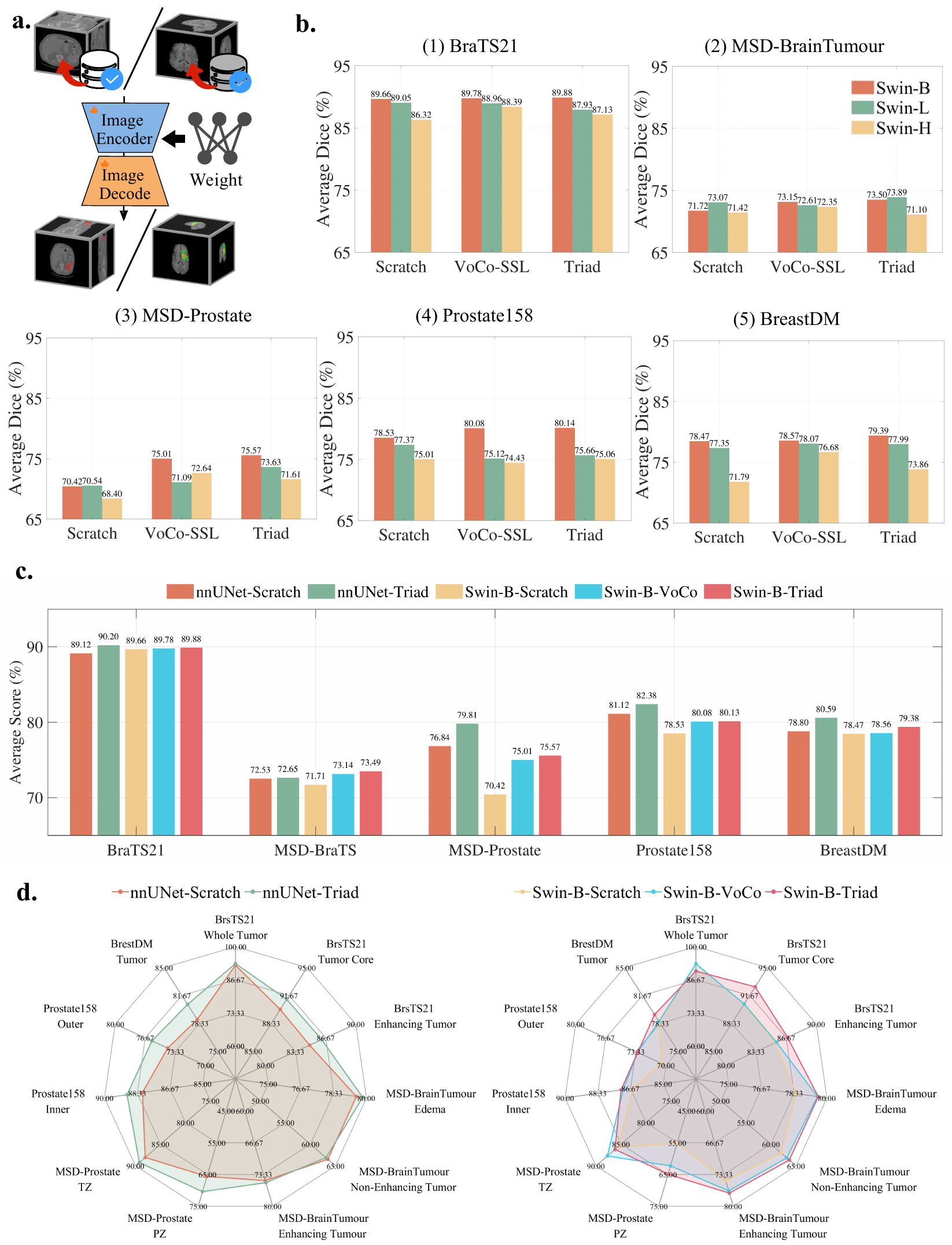}
	\caption{\textbf{Study on within-domain 3D tumor segmentation.} a. Image segmentation with encoder-decoder architecture by loading the weights of Triad. b. We compare the performance of Scratch, VoCo-SSL and Triad on 5 within-domain datasets based on 3 architectures: Swin-B/L/H. c. We select the nnUNet and Swin-Transformer-Base architectures, along with 3 different weight-loading strategies, and analyze their cross-effects on performance across 5 within-domain datasets. d. Consistent with the setting in subfig. c., the radar chart of each category shows the overall advantage of Triad in tumor segmentation. }
	\label{fig:seg_enc_dec} 
\end{figure}

We present results from 25 downstream datasets across three types of evaluation tasks and two data modality settings. These downstream tasks are categorized into within-domain and out-of-domain tasks.
Within-domain downstream tasks utilize the same data modalities and structures as those in the pre-training phase, including brain, breast, and prostate MRI. These tasks assess whether Triad has successfully learned structural and modality representations during pre-training, thereby improving performance on related tasks.
Conversely, out-of-domain downstream tasks involve data modalities or organs different from those in the pre-training stage, such as liver CT or atrial MRI. These tasks evaluate whether the knowledge acquired by Triad during pre-training can be effectively transferred to and applied in new modalities or structures.
Based on these two data modality settings, we evaluate three task types: 3D structure/tumor segmentation (Fig. \ref{fig:seg_enc_dec} and Fig. \ref{fig:seg_withoutdomain}), organ/cancer classification (Fig. \ref{fig:cls}), and 3D medical image registration (Fig. \ref{fig:registration}).

\subsection{3D organ/tumor segmentation}

We first evaluated the effectiveness of Triad on 3D organ/tumor segmentation. As shown in Fig. \ref{fig:seg_enc_dec} (a), we initialize the encoder with the parameters learned during pre-training, while the decoder is randomly initialized.
We evaluate Triad on 17 extensive 3D MRI and CT semantic segmentation datasets, including five MRI datasets for within-domain tasks: BraTS21\citep{baid2021rsna}, MSD\citep{antonelli2022medical}-BrainTumour, BreastDM\citep{zhao2023breastdm}, Prostate158\citep{adams2022prostate158}, and MSD-Prostate; and 12 datasets covering different organs or modalities for out-of-domain tasks: MM-WHS-MRI\citep{zhuang2018multivariate}, ATLAS-MRI\citep{quinton2023tumour}, Abdomen 1K\citep{ma2021abdomenct}, Kipa22\citep{he2021meta}, MSD-Pancreas, MSD-Liver, MSD-Heart, MSD-Hippocampus, MSD-Lung, MSD-HepaticVessel, MSD-Spleen, and MSD-Colon.
We use the Dice Similarity Coefficient (DSC) as the primary evaluation metric, consistent with public benchmarks.

\subsubsection{Influence of model parameter scale on model performance}

It is widely believed that increasing model parameters enhances the performance of downstream tasks in foundation models~\citep{oquab2023dinov2,chen2024towards,ghesu2022contrastive,amadou2024echoapex}. This trend has been observed in various domains, including natural images~\citep{oquab2023dinov2}, x-ray~\citep{ghesu2022contrastive}, and other medical imaging modalities\citep{chen2024towards,amadou2024echoapex}.
However, some studies have reported contradictory findings, particularly in 3D CT imaging\citep{wu2024voco} and vision language models~\citep{shi2025we,mei2024bigger}. In this study, we quantitatively analyze the scaling behavior of Triad pre-training. We conduct a series of experiments with varying model architectures to systematically evaluate their impact on performance.

We specifically select the widely used SwinUNETR architecture~\citep{he2023swinunetr}. The SwinUNETR encoder utilizes different variants of Swin Transformer~\citep{dosovitskiy2020image}, including Swin-B (Base), Swin-L (Large), and Swin-H (Huge).
We evaluate the impact of model parameter scaling on the within-domain 3D tumor segmentation task and compare it with Scratch and VoCo-SSL~\citep{wu2024voco}. VoCo-SSL is a vision foundation model pre-trained on 160K 3D CT scans using a self-supervised model distillation scheme and is considered state-of-the-art in 3D medical imaging. Scratch denotes training from scratch without using any pre-trained weights.

As shown in Fig. \ref{fig:seg_enc_dec} b (1) - b (5), the average DSC reported by Swin-B-Scratch across the five datasets is 77.76\%. In comparison, Swin-B-VoCo-SSL achieves an average DSC of 79.31\% (\textcolor{blue}{+1.55\%}), while Swin-B-Triad achieves 79.66\% (\textcolor{blue}{+1.90\%}).
These results indicate that pre-training the upstream encoder can markedly enhance the performance of downstream tasks. This finding is consistent with previous studies\citep{wu2024voco,amadou2024echoapex}. Notably, Swin-B-Triad outperforms Swin-B-VoCo-SSL by \textcolor{blue}{+0.35\%}, which can be attributed to Triad's use of MRI data for pre-training, whereas VoCo-SSL is pre-trained on CT data. This suggests that greater alignment between the data modalities of upstream and downstream models leads to improved downstream performance.
A key finding across 15 experiments comparing the Swin-B/L/H architectures is that 11 of these experiments indicate that increasing the size of model parameter does not consistently lead to performance improvements. This observation aligns with findings from VoCo-SSL\citep{wu2024voco}. A possible explanation is that excessive model parameters may lead to overfitting on small downstream datasets. Even with robust initial parameters from the upstream pre-trained model, downstream performance may still be adversely affected by overfitting.

\subsubsection{Within-domain 3D tumor segmentation}

The upstream 3D MRI data used for pre-training is derived from three organs: the brain, breast, and prostate. Consequently, the downstream within-domain 3D tumor segmentation task employs data from the same modality and organs to evaluate performance, assessing whether Triad has effectively learned structural and modality representations during pre-training.
Specifically, we select two brain datasets: BraTS21~\citep{baid2021rsna} and MSD-BrainTumour~\citep{antonelli2022medical}, one breast dataset: BreastDM~\citep{zhao2023breastdm}, and two prostate datasets: MSD-Prostate~\citep{antonelli2022medical} and Prostate158~\citep{adams2022prostate158}.
We employ two widely used network architectures: nnUNet~\cite{isensee2021nnu} and SwinUNETR (Swin-B).

The nnUNet-Scratch achieves an average DSC of 79.68\%, whereas nnUNet-Triad improved this to 81.13\%, marking an increase of \textcolor{blue}{1.45\%}. Considering the performance of the Swin-B encoder under the three initialization settings discussed earlier, it is evident that pre-trained parameters consistently outperform random initialization, regardless of the model structure (nnUNet or SwinUNETR) or the pre-training strategy (VoCo-SSL or Triad) employed.
Furthermore, the results indicate that nnUNet generally outperforms SwinUNETR in segmentation tasks. A radar chart presenting DSC for each category in Fig. \ref{fig:seg_enc_dec} (d). This chart demonstrates that Triad excels in segmenting fine-grained tumors.
For example, the BraTS21 Tumor Core represents the core region of the tumor, which serves as the primary therapeutic target and excludes the edema area. The BraTS21 Enhancing Tumor represents the actively invasive tumor and serves as a key indicator of tumor grade and recurrence. Notably, Triad outperforms Swin-B-VoCo by \textcolor{blue}{+2.08\%} and \textcolor{blue}{+1.38\%} in both categories.

\begin{figure}[!ht] 
	\centering
	\includegraphics[width=0.95\textwidth]{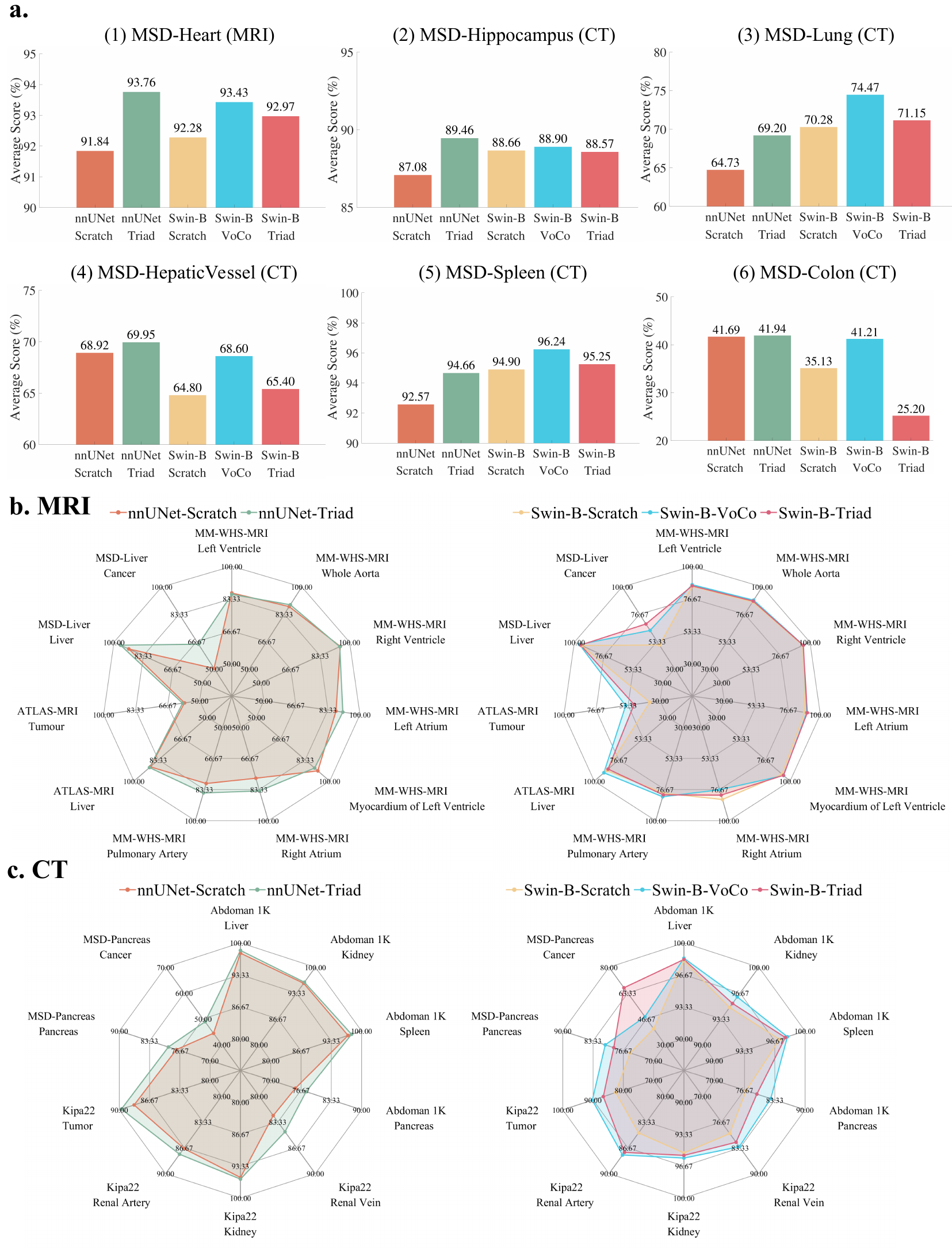}
	\caption{\textbf{Study on out-of-domain organ/tumor segmentation.} a. We select the nnUNet and Swin-Transformer-B architectures, along with three different weight loading strategies, and analyze their cross-effects on performance across six MSD CT datasets. b. Consistent with the setting of subfig. a., the radar chart shows the performance comparison of each category in MM-WHS-MRI, ATLAS-MRI, and MSD-Liver. c. Consistent with the setting of subfig. a., the radar chart shows the performance comparison of each category in Abdoman 1K, Kipa22, and MSD-Pancreas.}
	\label{fig:seg_withoutdomain} 
\end{figure}

\subsubsection{Out-of-domain organ/tumor segmentation}

We further assess whether the knowledge acquired by Triad during pre-training can be effectively transferred to and applied to other organs or a different imaging modality.
To achieve this, we select four MRI datasets from other organs: MSD-Heart, MSD-Liver, MM-WHS-MRI~\citep{zhuang2018multivariate}, and ATLAS-MRI~\citep{quinton2023tumour}. Additionally, we incorporated eight CT datasets: MSD-Hippocampus, MSD-Lung, MSD-Pancreas, MSD-HepaticVessel, MSD-Spleen, MSD-Colon, Abdomen 1K~\citep{ma2021abdomenct}, and Kipa22~\citep{he2021meta}.
We continue to use the nnUNet and SwinUNETR (Swin-B) architectures and employ three parameter initialization methods: training from scratch, VoCo-SSL, and Triad.

As shown in Fig. \ref{fig:seg_withoutdomain} (a), an interesting observation is that, when using the Swin-B architecture, the three initialization methods rank in performance as follows: VoCo (77.14\%) > Scratch (74.34\%) > Triad (73.09\%)\footnote{We report the average DSC of experiments (1)-(6).}.
We also observed that VoCo-SSL was pre-trained on CT data containing more than 16 organs or tumors, covering the organs used in experiments (1)-(6). We believe this is the primary reason for its superior performance.
Therefore, we conclude that \textbf{pre-training maximizes performance when the data modality and organ of upstream and downstream tasks are consistent.}
Nevertheless, compared to both VoCo-SSL and training from scratch, nnUNet-Triad achieved an improvement of \textcolor{blue}{+2.02\%}, demonstrating that Triad can effectively generalize to other data modalities and organs.
Fig. \ref{fig:seg_withoutdomain} (b) and (c) present radar charts for each class on both CT and MRI datasets. Notably, marked improvements are observed in tumor segmentation rather than organs.
For example, on MSD-Liver Cancer, nnUNet-Triad outperforms nnUNet-Scratch by \textcolor{blue}{+14.77\%}, while Swin-B-Triad surpasses Swin-B-Scratch by \textcolor{blue}{+17.97\%}.

\subsection{Organ/cancer classification}

We next evaluate the performance of Triad on organ and cancer classification tasks.
As shown in Fig. \ref{fig:cls} (a), we initialize the encoder with the parameters learned during pre-training, apply an average pooling operation to the output of its final layer, and pass the resulting features through a two-layer linear classifier to predict the probability distribution of the categories.
We evaluate Triad on five widely recognized 3D CT and MRI classification datasets, including two MRI datasets for within-domain classification: ADNI~\cite{jack2008alzheimer} and BreastDM~\cite{zhao2023breastdm}; two CT datasets for out-of-domain classification: OrganMNIST3D~\cite{yang2023medmnist} and LUNA16~\cite{setio2017validation}; and one additional MRI dataset for out-of-domain classification: LLD-MMRI~\cite{lou2024sdrformer}.
We use classification accuracy (Acc) as the primary evaluation metric.

\subsubsection{Within-domain organ/cancer classification}

\begin{figure}[!ht] 
	\centering
	\includegraphics[width=0.99\textwidth]{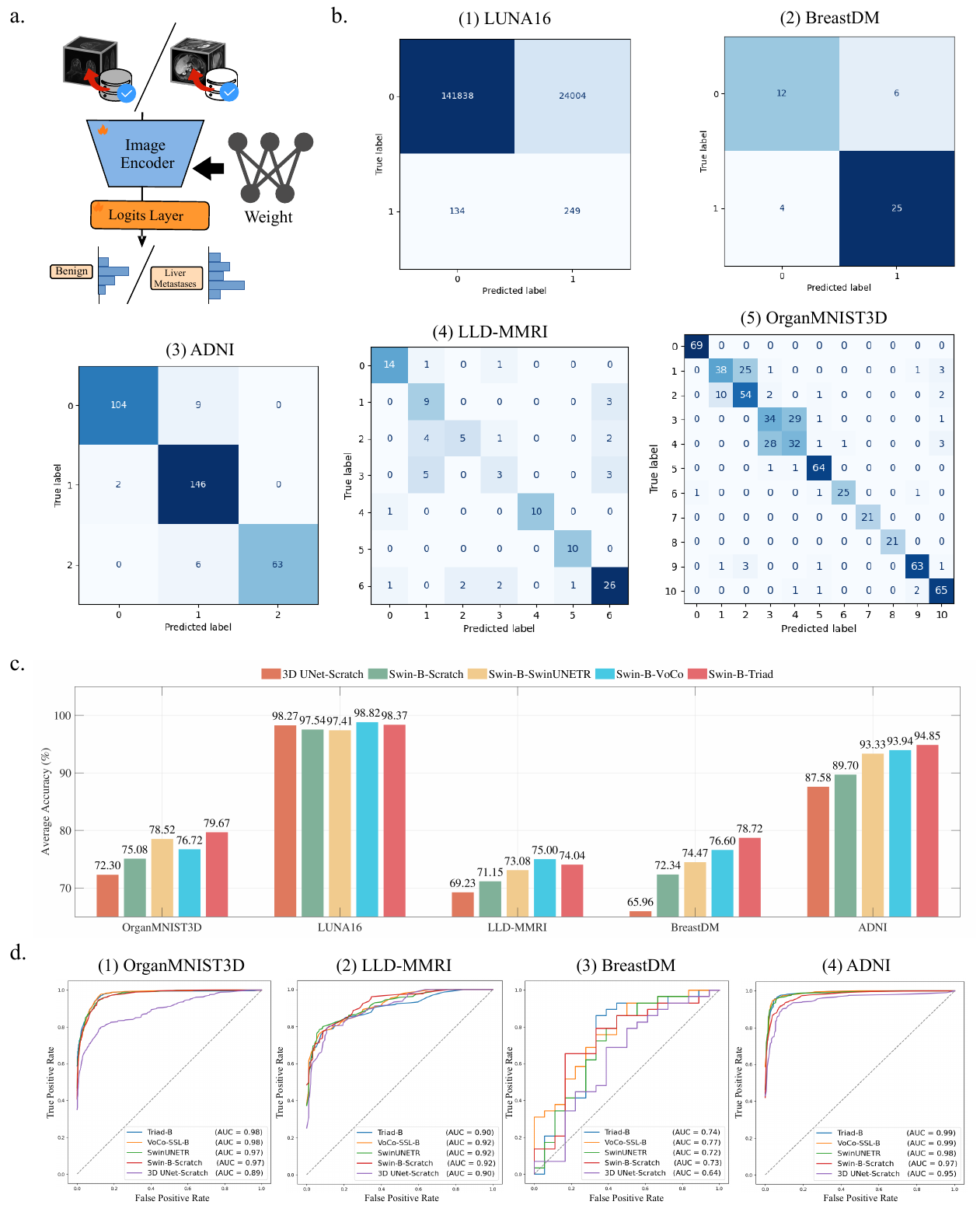}
	\caption{\textbf{Study on organ/cancer classification.} a. We use an encoder loaded with Triad weights and a two-layer linear classifier for classification tasks. b. Confusion matrices of the 5 datasets when using Swin-B-Triad as the encoder. The meaning of each category number is shown in Table \ref{table:t2}. c. We select the 3D UNet and Swin-Transformer-Base architectures, along with 3 different weight loading strategies, and analyze their cross-effects on performance across 5 CT/MRI datasets. d. Consistent with the setting of subfig. c., we plot the ROC curve of each scheme on 4 datasets.}
	\label{fig:cls} 
\end{figure}

As shown in Fig. \ref{fig:cls} (c), we compare two architectures, 3D UNet and Swin-B, using four initialization methods: training from scratch, SwinUNETR~\cite{cao2022swin}, VoCo-SSL, and Triad.
We observe that on both the ADNI and BreastDM datasets, Swin-B-Scratch achieves an average accuracy that is \textcolor{blue}{+4.25\%} higher than 3D UNet-Scratch. A similar trend is observed in the LLD-MMRI and OrganMNIST3D datasets.
The only exception is the LUNA16 dataset, where 3D UNet-Scratch achieves an accuracy that is \textcolor{blue}{+0.73\%} higher than Swin-B-Scratch.
These findings provide strong evidence that the Swin-B architecture is better suited for classification tasks.

Next, we compare the impact of three different pre-trained models on downstream performance.
SwinUNETR is pre-trained on approximately 5K CT volumes, whereas VoCo-SSL utilizes 160K CT volumes. According to the reported accuracy, VoCo-SSL achieves an average accuracy that is \textcolor{blue}{+1.37\%} higher than SwinUNETR.
Triad is pre-trained on 131K MR volumes and achieves an average accuracy that is \textcolor{blue}{+1.52\%} higher than VoCo-SSL.
These results indicate that both the modality and scale of pre-training data positively impact downstream performance.

\begin{figure}[!t] 
	\centering
	\includegraphics[width=0.81\textwidth]{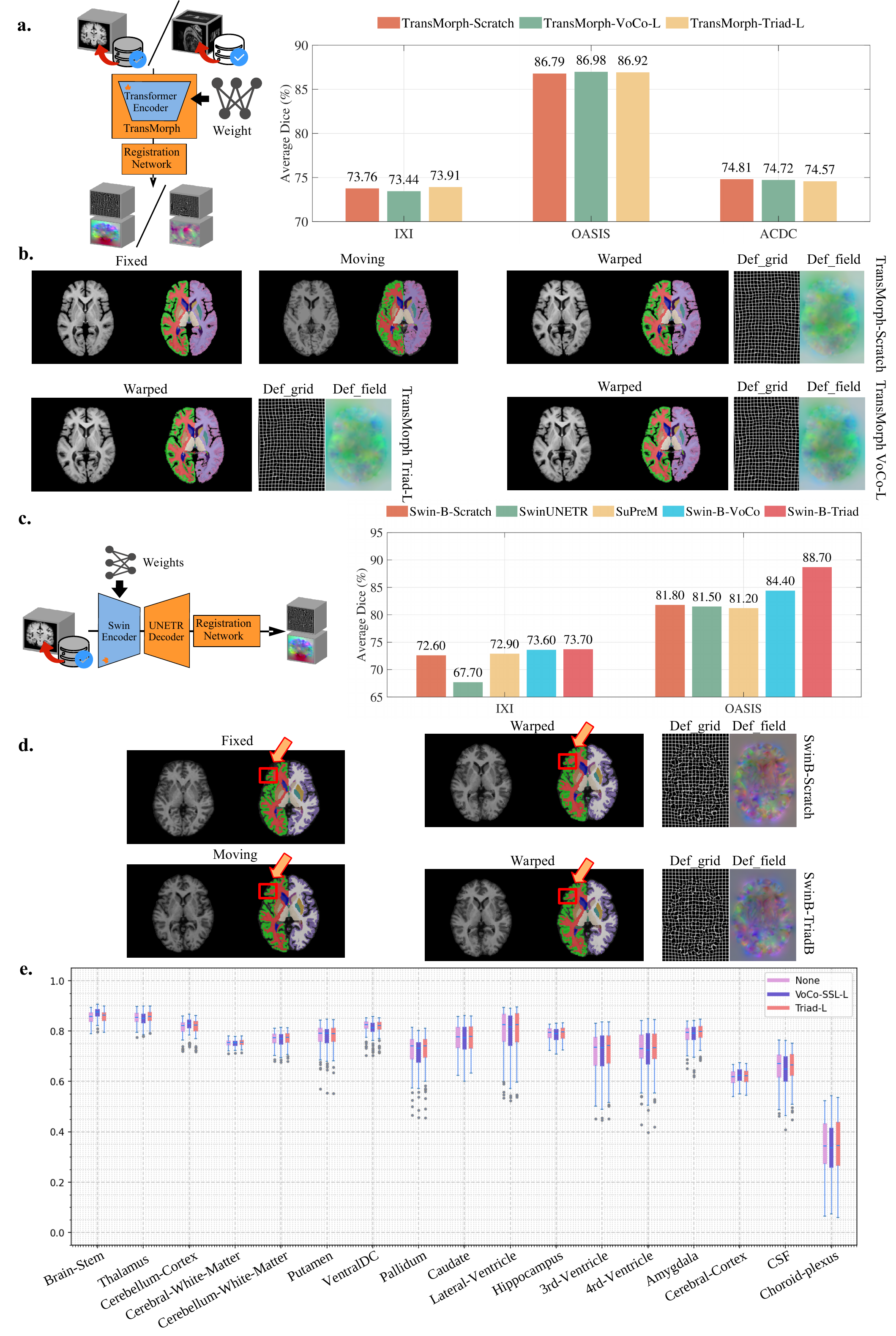}
	\caption{\textbf{Study on 3D medical image registration.} a. We adopt the TransMorph architecture, use Swin-Transformer Large as the encoder, and load the pre-trained weights of Scratch, Triad, and VoCo-SSL for regristation task. The bar chart on the right shows the average dice scores of the 3 weight loading methods on the 3 datasets. b. Under the setting of subfig. a., the visualization results of various registration methods in the IXI dataset. c. We adopt the SwinUNETR architecture, use Swin-Transformer Base as the encoder. The bar chart on the right shows the average dice scores of the 5 weight loading methods on the 2 datasets. d. Under the setting of subfig. c., the visualization results of various registration methods in the OASIS dataset. e. Under the setting of subfig. a., boxplots with Dice scores of various registration methods in the abdomen IXI dataset.}
	\label{fig:registration} 
\end{figure}

\subsubsection{Out-of-domain organ/cancer classification}
As illustrated in Fig. \ref{fig:cls} (c), Triad achieves the highest performance in the organ classification task and ranks second in both lung nodule and liver lesion classification tasks.
Notably, Triad still outperforms training from scratch by \textcolor{blue}{+1.02\%}, demonstrating its effectiveness in generalizing across diverse imaging modalities and organ types.
Furthermore, we provide the confusion matrix~\cite{heydarian2022mlcm} for Swin-B-Triad across the five datasets.
Fig. \ref{fig:cls} b (1) shows that when Swin-B-Triad is applied to an out-of-domain classification task with a highly imbalanced category distribution, the model struggles to classify minority classes accurately.
In the OrganMNIST3D classification task, Triad fails to distinguish categories 1–4 accurately.
These findings suggest that while pre-trained parameters enhance overall downstream performance, addressing challenges such as data imbalance and hard example mining may require specialized sampling strategies or model architectures.
Additionally, we present the ROC curves for four datasets in Fig. \ref{fig:cls} (d).
The ROC curves of all pre-trained models exhibit significant overlap, whereas models trained from scratch show markedly inferior performance, particularly on OrganMNIST3D and ADNI.

\subsection{3D medical image registration}

Finally, we evaluate the performance of Triad on the 3D medical image registration task.
As illustrated in Fig. \ref{fig:registration} (a) and (c), we explore two different parameter initialization strategies.
In Fig. \ref{fig:registration} (a), we employ the TransMorph~\citep{chen2022transmorph} architecture with a Swin-Transformer-L encoder, initializing it with pre-trained weights from Triad and VoCo-SSL.
In Fig. \ref{fig:registration} (c), we use the Swin-UNETR~\citep{hatamizadeh2021swin} architecture with a Swin-Transformer-B encoder, initializing it with pre-trained weights from Triad, VoCo-SSL, SuPreM, and SwinUNETR.
The decoder remains unchanged from the original method and is randomly initialized.
We evaluate Triad on three widely recognized 3D MRI registration datasets, including two brain datasets for within-domain registration: IXI~\cite{Brain-Development_IXI_2019} and OASIS~\cite{krentzel2023clem}, as well as one cardiac dataset for out-of-domain registration: ACDC~\cite{bernard2018deep}.
We use the Dice Similarity Coefficient (DSC) as the primary evaluation metric and report the best results after fine-tuning for 200 epochs.

\subsubsection{Comparison of pre-training strategies in TransMorph and SwinUNETR for 3D medical image registration}

Fig. \ref{fig:registration} (a) illustrates a bar chart depicting the DSCs for each dataset. TransMorph-Scratch achieves average DSCs of 73.76\%, 86.79\%, and 74.81\% on IXI, OASIS, and ACDC, respectively.
When employed Triad pre-trained weights, the DSCs are 73.91\% (\textcolor{blue}{+0.15\%}) on IXI and 86.92\% (\textcolor{blue}{+0.13\%}) on OASIS, but decrease to 74.57\% (\textcolor{red}{-0.24\%}) on ACDC.
Similarly, VoCo-SSL pre-training results in DSCs of 73.44\% (\textcolor{red}{-0.32\%}), 86.98\% (\textcolor{blue}{+0.19\%}), and 74.72\% (\textcolor{red}{-0.09\%}) on IXI, OASIS, and ACDC, respectively.
The performance of TransMorph under these three initialization strategies indicates that pre-trained parameters do not consistently yield improvements over random initialization in 3D medical image registration.
This observation applies to both within-domain (IXI, OASIS) and out-of-domain (ACDC) tasks, with some cases even exhibiting marginal performance declines.

Fig. \ref{fig:registration} (c) presents DSC performance under the SwinUNETR architecture, comparing five initialization strategies: Scratch, SwinUNETR, SuPreM, VoCo-SSL, and Triad.
Specifically, Swin-B-Scratch achieves DSCs of 72.60\% on IXI and 81.80\% on OASIS.
Swin-B-VoCo achieves DSC scores of 73.60\% (\textcolor{blue}{+1.00\%}) on IXI and 84.40\% (\textcolor{blue}{+2.60\%}) on OASIS.
Notably, Triad proves to be the most effective pre-training method, achieving DSC scores of 73.70\% (\textcolor{blue}{+1.10\%}) on IXI and 88.70\% (\textcolor{blue}{+6.90\%}) on OASIS.
These improvements are particularly pronounced on the OASIS dataset, where Triad outperforms other initialization methods by a significant margin.

By integrating these findings with the TransMorph results in Fig. \ref{fig:registration} (a), we observe that partially loading encoder weights, as done in TransMorph, while randomly initializing the remaining parameters, may introduce inconsistencies, potentially limiting the benefits of pre-training.
In contrast, SwinUNETR employs a fully pre-trained encoder, thereby eliminating random initialization in that module. This allows the network to leverage pre-trained features more effectively, leading to substantial improvements in 3D medical image registration.

\subsubsection{Impact of initialization method on registration performance}

Fig. \ref{fig:registration} (b) presents the visualization results for the IXI dataset using the TransMorph architecture.
Regardless of the initialization method used for fine-tuning, the observed  improvements in the mask appear similar, with no substantial enhancements detected.
In contrast, Fig. \ref{fig:registration} (d) presents the visualization of the OASIS dataset using the SwinUNETR architecture, where improvements in the mask (indicated by the red arrow) are noticeable.
These visualizations provide intuitive evidence supporting our previous assertion that incomplete pretraining initialization of the encoder may lead to model confusion.

Furthermore, Fig. \ref{fig:registration} (e) depicts the DSC distributions for the IXI dataset.
SwinUNETR initialized with Triad weights achieves the highest registration performance across most organs, including the thalamus, cerebral white matter, cerebellar white matter, pallidum, caudate, lateral ventricle, hippocampus, third ventricle, fourth ventricle, and amygdala.
Notably, this superior performance can be attributed to the inclusion of abundant 3D MRI brain organ data in the upstream pretraining tasks.
Regardless of the initialization weights, the DSC and registration performance for the choroid plexus remains low, likely due to its complex anatomical attachments to surrounding structures and its diffuse morphological characteristics.

\section{Discussion}
\label{sec:discussion}


In this study, we constructed a large-scale 3D MRI dataset, known as TriadMR-131K, which consists of 131,170 volumes from 19,721 patients across 36 clinical datasets.
This extensive collection includes a diverse collection of 3D MRI data from three organs, including the brain, breast, and prostate. It includes modalities such as T1-w, T2-w, FLAIR, DWI-MRI, fMRI, and DCE-MRI.
Additionally, we extract imaging descriptions from the metadata of each 3D volume. These descriptions detail the imaging modality and associated device parameters.
Using this dataset, we develop Triad, a vision foundation model tailored for 3D MRI.
Triad employs widely used autoencoder architecture to learn robust representations and incorporates organ-independent imaging descriptions to constrain the semantic distribution of the visual modality.
We evaluate Triad on three tasks, including organ and tumor segmentation, organ and cancer classification, and medical image registration. These tasks are assessed across two data modalities---within-domain and out-of-domain--- across 25 downstream datasets.
By initializing models with Triad's pre-trained weights, nnUNet-Triad improves segmentation performance by 2.51\% over nnUNet-Scratch across 17 datasets. Swin-B-Triad achieves a 3.97\% improvement over Swin-B-Scratch in classification tasks across five datasets. SwinUNETR-Triad improves by 4.00\% compared to SwinUNETR-Scratch in registration tasks across two datasets.
Triad outperforms baseline models across all downstream tasks and exceeds existing state-of-the-art models in most cases.
Overall, Triad's seamless adaptability across downstream tasks highlights its potential as a versatile and efficient tool for diverse clinical applications.
This adaptability paves the way for enhanced diagnostic accuracy in 3D MRI.

Our findings highlight several key observations regarding the effectiveness of pretraining:

\begin{itemize}

\item Most vision foundation models for medical image analysis have been pre-trained on large-scale 3D CT datasets~\citep{tang2022self,ye2024continual,zhuang2024mim,jiang2023anatomical}, highlighting the prevalence of CT-based self-supervised learning in this domain.
However, our experimental results indicate that pre-training on modality-specific upstream data is more effective. This is demonstrated by the superior performance of Triad, which was pre-trained on MRI data, in 3D MRI-based segmentation, classification, and registration tasks.
This finding underscores a critical gap in current research, namely the absence of large-scale 3D MRI pre-training datasets for foundation models.
By assembling a diverse collection of 3D MRI data, our work addresses this gap and establishes a more appropriate pre-training paradigm for 3D MRI-based medical imaging tasks.
Furthermore, although this study employs an autoencoder (AE) architecture that is not the most advanced in self-supervised learning, our comparison with VoCo-SSL, a state-of-the-art distilled model, suggests that even a relatively simple AE-based framework can outperform more complex models under certain conditions.
This finding highlights an important insight: data quality and task-specific alignment are more critical than model complexity, suggesting that the future of medical AI should emphasize comprehensive and representative pre-training datasets rather than focusing solely on architectural advancements.

\item The experimental results of Triad pre-training highlight the dual impact of model parameter scale and task alignment.
In the context of 3D organ and tumor segmentation, pre-trained encoders consistently outperform their randomly initialized counterparts across various architectures.
However, increasing the scale of the model parameters does not consistently lead to improved performance, as evidenced by 11 out of 15 experiments showing no marked gain.
This suggests that while larger models theoretically have greater representational capacity, the limited size of downstream medical imaging datasets may lead to overfitting, offsetting the benefits of pre-training.
Furthermore, Triad, which is pre-trained on MRI data, surpasses the CT-based VoCo-SSL model in MRI-related tasks, underscoring the critical role of modality alignment in transfer learning for medical imaging.
Notably, Triad demonstrates superior segmentation performance in fine-grained tumor regions, such as the enhancing tumor component in BraTS21. This suggests that modality alignment not only affects overall segmentation accuracy but also improves the precise delineation of clinically relevant tumor subregions.

\item In out-of-domain tasks, Triad exhibits strong performance in MRI-based applications but lags behind VoCo-SSL in CT-based segmentation.
Although Triad does not surpass VoCo-SSL in CT tasks, it significantly outperforms models trained from scratch and achieves notable improvements in certain tumor segmentation tasks, such as liver cancer.
These results suggest that the benefits of pre-training are more pronounced for lesion recognition than for anatomical organ segmentation.
Additionally, in classification tasks, Swin-B consistently outperforms 3D UNet, suggesting that Transformer-based architectures may be more suitable for medical image classification.
However, in highly imbalanced datasets, Triad-pretrained models struggle with minority class discrimination, underscoring the limitations of pre-training alone in mitigating data imbalance.
These findings suggest that future work should incorporate advanced sampling strategies or hard example mining techniques to further improve model generalization across diverse medical imaging tasks.

\item In registration tasks using TransMorph, loading pre-trained weights from Triad and VoCo-SSL resulted in only marginal improvements in within-domain registration tasks, such as IXI and OASIS. However, in some cases, including the out-of-domain ACDC dataset, performance slightly declined compared to random initialization.
These results suggest that partially loading encoder weights, as in TransMorph, may introduce inconsistencies that impede optimal feature learning during fine-tuning.
In contrast, SwinUNETR, which employs a fully pre-trained encoder, achieved more substantial improvements when initialized with Triad weights.
This effect was particularly evident in the OASIS dataset, where the DSC increased markedly by 6.90\% compared to random initialization.
These findings indicate that leveraging a fully pre-trained encoder enables more effective feature transfer, thereby improving registration accuracy.
Furthermore, the superior performance of Triad over other pretraining approaches underscores the advantages of task-specific pretraining strategies tailored for 3D MRI data.

\end{itemize}

Despite the strengths of our study, there are several limitations in both our model and methodology.
First, the text data used for pre-training in Triad-131K primarily focuses on device parameters and imaging modalities rather than image content. This restricts the extension of the single-modality vision foundation model into a vision-language foundation model.
Second, the computational demands of pre-training such a large-scale 3D MRI dataset prevent us from employing ensemble methods and fixed-step validation, both of which could further enhance the performance of downstream tasks.
We retain model parameters only at iteration 20,000.
Third, fine-tuning across a broad range of downstream tasks is highly resource-intensive in terms of manpower, computational cost, and time.
Following VoCo’s approach, we report results solely on fold 0 instead of conducting multi-fold cross-validation, which may introduce variability due to data distribution effects.
Fourth, additional exploration is required to optimize Triad and enhance its performance on downstream tasks.
For instance, in the registration task, we initialize TransMorph with Triad’s pre-trained weights and subsequently fine-tune the model.
The training process involves balancing multiple loss functions; however, we use default parameters without exploring a broader optimization space.
Future research should explore more effective parameter configurations and optimization strategies.

Future research can focus on the following aspects. First, improving data quality.
Despite having implemented various preprocessing pipelines, the presence of low-quality cases remains a challenge.
Numerous studies have emphasized the crucial role of data quality in pre-training~\citep{yang2024freemask,kirillov2023segment}.
In future work, we plan to design robust automated screening pipelines to screen each volume and enhance data quality.
Second, assigning structured electronic health reports to each 3D MRI volume in the Triad-131K dataset to facilitate pre-training for vision-language foundation models.
Additionally, expanding downstream tasks to include assistive report generation, visual question answering, and cross-modal medical image retrieval.
Lastly, the current pre-training strategy is not limited to MRI scans.
Moving forward, we plan to integrate Triad-131K with the largest available CT, PET, X-ray, and ultrasound datasets to pre-train foundation models that generalize effectively across a broad range of clinical tasks, rather than being confined to specific imaging modalities.

\section{Methods}

The following sections are structured are organized to provide a comprehensive overview of our methodologies and findings. Initially, we will introduce the data curation process for TriadMR-131K, as well as the protocol established for Triad pre-training. Following this, we will articulate the implementation strategies employed in three distinct experimental paradigms: 3D organ and tumor Segmentation, organ and cancer classification, and 3D medical image registration. 

\subsection{Pretraining}
\label{sec:data}

\noindent \textbf{Curation of TriadMR-131K dataset.} To ensure the quality and diversity of data for model pretraining, we curated a large-scale 3D MRI dataset of 131,170 3D volumes derived from 19,721 patients across 36 clinical datasets. TriadMR-131K comprises a diverse collection of 3D MRI data spanning three organs (brain, breast, and prostate), featuring modalities such as T1-w, T2-w, FLAIR, DWI-MRI, fMRI, DCE-MRI.
To standardize all sub-datasets, we used the same preprocessing protocol for all organs: we used the dicom2nifti\footnote{\url{https://github.com/icometrix/dicom2nifti}} package to convert all DICOM-format 2D slice collections into NIfTI-format 3D volumes. For 4D volume data, such as DCE-MRI, we took the $\frac{(t-1)}{2}$th or $\frac{(t+1)}{2}$th 3D slice to replace the original 4D data, where $t$ denotes the $t$th time step. All corrupted volumes were deleted during the conversion process.
Next, we reformatted all MRI scans so that the first axis points from left to right, the second from posterior to anterior, and the third from inferior to superior. We then resampled the images to a 1 mm resolution using bilinear interpolation. We also resized all images to (256,256,128) using trilinear interpolation.
To save storage space, we stored most of the 3D volume data types as UINT16 and the rest as Float32.
Finally, the brain MRI data involved 51,112 series from 37,436 examinations of 12,994 patients; the breast MRI data involved 46,116 series from 8,180 examinations of 3,834 patients; the prostate MRI data involved 33,942 series from 9,941 examinations of 4,639 patients. The statistical information of the 36 datasets is shown in Table \ref{table:t1}. Note that due to deletions during the conversion process, the numbers in the table are usually lower than the officially published figures.
In addition, we extracted the imaging description from the metadata of each 3D volume, which describes the imaging modality and related device parameters. Since it is organ-independent information, it helps to adjust the positional relationship of each modality in the semantic space, thereby improving the generalization ability of the model.
We tried to avoid any overlap between the datasets used in pre-training and all downstream evaluation sets to minimize the risk of data contamination.

\noindent \textbf{Protocol of Triad pre-training.} Triad uses nnUNet (31M)\cite{isensee2021nnu} and SwinTransformer\cite{hatamizadeh2021swin} for the image encoder. Furthermore, SwinTransformer is expanded into Swin-B (72.8M), Swin-L (290.4M), and Swin-H (11.6B) according to feature sizes of 48, 96, and 192 to study the parameter scaling law (Fig. 2b).
We use GTR-T5-Large~\cite{ni2021large} as the text encoder instead of CLIP~\cite{radford2021learning} because the text embedding of T5 can provide a more semantically nuanced distribution and is suitable as a supervisory signal to guide the alignment of visual modality distributions to the semantic space, rather than just as a representation of relative similarity in contrastive learning. This means that the parameters of the text encoder are frozen and do not participate in gradient updates.
In the pretraining stage, we use an upsampling 3D CNN as the image decoder for the self-supervised reconstruction task. In the downstream task, we only keep the parameters of the image encoder and replace the image decoder with the task adapter.
Triad's pretraining uses L1 loss as the reconstruction loss and log-ratio loss~\cite{kim2019deep} to align the distribution of the visual modality with that of the textual modality. To prevent the log-ratio loss from dominating the optimization process, its weight is set to 0.01.
The architectural hyperparameters of the models involved are shown in Table \ref{table:t3}.

\subsection{3D organ/tumor segmentation}
\label{sec:sts}

\subsubsection{Curation of segmentation datasets}

\textbf{Within-domain segmentation datasets}. The BraTS21 dataset~\citep{baid2021rsna}, released as part of the 2021 RSNA-ASNR-MICCAI Brain Tumor Segmentation Challenge, consists of multi-institutional, multi-modal MRI scans (T1-w, T1 postcontrast, T2-w, FLAIR) from patients with glioblastoma or lower-grade glioma. Each case includes expert annotations delineating tumor subregions (enhancing tumor, edema, and necrotic core). The core challenge portion provides 1,251 labeled scans for training.
The MSD–BrainTumour dataset~\citep{antonelli2022medical} (Task 01 in the Medical Segmentation Decathlon) includes 484 preoperative multi-modal MRI scans (T1-w, T1 postcontrast, T2-w, FLAIR) sourced primarily from earlier BraTS collections.
For BreastDM~\citep{zhao2023breastdm}, the original publication reports a new breast MRI dataset (with dynamic contrast-enhanced volumes) consisting of 232 scans.
The Prostate158 dataset~\citep{adams2022prostate158} offers 158 MRI scans (T2, ADC, DFI) with detailed prostate annotations.
Finally, the MSD–Prostate dataset~\citep{antonelli2022medical} (Task 05 in the MSD) contains 32 T2-w, ADC map, and DFI scans with corresponding prostate zonal annotations (central gland and peripheral zone).

\textbf{Out-of-domain segmentation datasets}. The MM-WHS–MRI subset~\citep{zhuang2018multivariate} from the Multimodality Whole Heart Segmentation challenge consists of around 20 annotated 3D MRI volumes of the heart.
ATLAS–MRI~\citep{quinton2023tumour} is a publicly available dataset of contrast-enhanced MRI for hepatocellular carcinoma (HCC), which consists of 60 scans.
The Abdomen-1K~\citep{ma2021abdomenct}, released as part of the MICCAI FLARE 2022 Playground subtask 1, includes a training set adapted from MSD Pancreas (281 cases) and NIH Pancreas (80 cases), where all 361 CT scans are from the portal phase.
Kipa22~\citep{he2021meta} comes from the Kidney PArsing Challenge 2022, and its goal is to segment 3D kidneys, renal tumors, arteries, and veins.  It  released 70 training sets with detailed annotations.
Lastly, the remaining MSD tasks, namely MSD-Pancreas (281 training scans), MSD-Liver (131 training scans), MSD-Heart (20 training scans), MSD-Hippocampus (260 training scans), MSD-Lung (63 training scans), MSD-HepaticVessel (303 training scans), MSD-Spleen (41 training scans), and MSD-Colon (126 training scans), do not provide official validation sets.
For all the above datasets, we keep the same split method as provided by VoCo~\citep{wu2024voco}.

\subsubsection{Fully supervised finetuning with nnUNet framework}

In our image segmentation experiments, we adopt nnUNetv2 as a unified framework to ensure consistent data preprocessing for fair comparisons across different models. Within this framework, we have implemented Swin-Transformer Base/Large/Huge networks, thereby aligning the training protocols.
For each publicly available dataset with detailed annotations, nnUNetv2's built-in code is used to perform a 5-fold cross-validation split. Note that, in order to make a fair comparison with VoCo, we report the 0th fold in the 5-fold cross-validation.
Throughout training, nnUNet models are trained for 300 epochs, while Swin-Transformer models are trained for 150 epochs, and we select the model with the highest validation performance for final evaluation.
We employ an SGD optimizer with an initial learning rate of 0.01, following nnUNet's default decay schedule. VoCo-SSL pretrained weights are sourced from the code library\footnote{\url{https://github.com/Luffy03/Large-Scale-Medical}}.
Because our experimental setup closely matches that of VoCo, some of the results reported here are derived from the extended version of the original publication~\citep{wu2024voco}.

\subsection{Organ/cancer classification}
\label{sec:occ}
\subsubsection{Curation of classification datasets}
\textbf{Within-domain classification datasets.} The ADNI dataset~\citep{jack2008alzheimer} (Alzheimer’s Disease Neuroimaging Initiative) is a longitudinal, multi-center, observational study that includes thousands of participants, from cognitively normal (CN) to those with mild cognitive impairment (MCI) or Alzheimer’s disease (AD).
In this study, we use a dataset consisting of participants who have screening, 6-month, 1-year, 18-month (MCI only), 2-year, and 3-year (normal and MCI only) scans, which is called ``ADNI1\_Complete 3Yr 1.5T,'' totaling 2,182 samples. Consistent with the literature~\cite{majee2024enhancing}, the training set, validation set, and test set contain 1,526, 326, and 330 samples, respectively.
We use NPPY~\cite{he2023neural} and its available pre-trained weights to convert raw MRI scans into uniformly sized skull-stripped, intensity-normalized brain volumes in standard coordinate space, and then reshape them into a smaller dimension of 96×96×96.
The BreastDM dataset~\citep{zhao2023breastdm} provides dynamic contrast-enhanced (DCE) breast MRI scans for lesion classification, containing 85 benign samples and 147 malignant samples. We adopt the same dataset split scheme as in the segmentation task.

\textbf{Out-of-domain classification datasets}. The OrganMNIST3D dataset~\citep{yang2023medmnist} is part of MedMNIST v2 and contains more than 1,700 3D CT volumes of 11 organs for classification. Its official distribution includes dedicated training sets (972 volumes), validation sets (161 volumes), and test sets (610 volumes). The samples in OrganMNIST3D are available in 28×28×28 and 64×64×64 versions, and we use the latter for evaluation.
The LUNA16 dataset~\citep{setio2017validation}, derived from the LIDC/IDRI collection, contains 888 thoracic CT scans for lung nodule analysis. LUNA16 includes a total of 551,065 candidate nodules, of which 1,120 nodules are detected as positive, represented by 1, and the rest are represented by 0. The full dataset is divided into 10 subsets, and we use subsets 0-5 as training sets, subset 6 as the validation set, and subsets 7-9 as test sets.
Finally, the LLD-MMRI dataset~\citep{lou2024sdrformer} contains 498 annotated multi-stage liver lesions from the same number of patients. The lesions are classified into seven categories: hepatocellular carcinoma (HCC), intrahepatic cholangiocarcinoma (ICC), hepatic metastasis (HM), hepatic cyst (HC), hepatic hemangioma (HH), focal nodular hyperplasia (FNH), and hepatic abscess (HA). The dataset has been pre-partitioned into a training set (316 lesions), a validation set (78 lesions), and a test set (104 lesions).

\subsubsection{Fully supervised finetuning with linear classifier}

As shown in Fig. \ref{fig:cls} (a), we use the parameters saved in the pretraining phase as the initial parameters of the encoder, perform an average pooling operation on the output of the last layer of the encoder, and then input it into a two-layer linear classifier to predict the probability distribution of the category.
In classification experiments, we set the ADNI dataset input size to 96×96×96, while all other datasets are resized to 64×64×64. Classifiers based on the Swin-Transformer are trained for 150 epochs, and those based on the 3D UNet are trained for 300 epochs. In each experiment, we report the best result.
We employ a learning rate of 1e-3, with the Adam optimizer, following a cosine decay schedule. Additionally, the first 5 epochs are used for warmup to stabilize training.

\subsection{3D medical image registration}
\subsubsection{Curation of registration datasets}

\textbf{Within-domain registration datasets}. The IXI dataset~\citep{Brain-Development_IXI_2019} consists of over 576 T1-weighted brain MRI scans from healthy volunteers collected at three different hospitals in London. Following the TransMorph~\citep{chen2022transmorph} protocol, we use 403 scans as the training set, 58 as the validation set, and 115 as the test set. The volumes are cropped to 160×192×224. Thirty annotated structures were used for evaluation.
The OASIS dataset~\citep{krentzel2023clem} (Open Access Series of Imaging Studies) includes 413 T1-weighted brain MRI scans from participants aged 18 to 96, with both healthy controls and patients exhibiting mild to moderate cognitive impairment. The original MR volumes are preprocessed using FreeSurfer~\cite{fischl2012freesurfer}, which includes spatial normalization, skull stripping, affine transformations, and automatic structural segmentation.
Following the TransMorph~\citep{chen2022transmorph} protocol, we use 394 scans as the training set and 19 scans as the validation set. Since there is no test set available, we employed the validation set for evaluation. The volumes are cropped to 160×192×224. 35 structures are used as ground truths to evaluate the performance.

\textbf{Out-of-domain registration datasets}. The ACDC dataset \citep{bernard2018deep} (Automated Cardiac Diagnosis Challenge) comprises 150 cardiac MRI scans in short-axis view, covering subjects with various heart conditions. The original challenge reserves 100 scans for training and 50 for testing. The volumes are cropped to 160×192×224. 

\subsubsection{Fine-tuning TransMorph/SwinUNETR for image registration}

For fine-tuning of TransMorph, we replaced the ``Transformer Encoder'' in the original framework with our own Swin-Transformer Encoder, loading the weights of Triad-L. The rest of the components, such as ``CNN Decoder,'' ``Affine Network,'' and ``Spatial Transform,'' are randomly initialized.
For fine-tuning of SwinUNETR, we load the weights of Triad-B into the encoder and randomly initialize the UNETR decoder. The pre-trained weights of SwinUNETR and SupreM are obtained from the code repository provided by VoCo.
Due to limited resources, we only fine-tune for 200 epochs on each set of experiments and select the best-performing results for reporting. The Adam optimizer is used for fine-tuning, and the batch size was 1. The learning rates for OASIS and IXI are 0.00005, while the learning rate for ACDC is 0.0001.
The remaining parameters, such as the type of loss function and weight factor, remain consistent with the default settings in the code provided by TransMorph.

\subsection{Computing hardware and software}
\label{sec:code}
We use pydicom 3.0.1 and dicom2nifti 2.5.0 for 2D slice sequences and 3D volume data preprocessing. We use Python 3.10.13 for all experiments and analyses in the study.
For the pretraining stage, we use the AdamW~\cite{loshchilov2017decoupled} optimizer with an initial learning rate of 1e-6, coupled with a cosine learning rate scheduler. The learning rate decays to zero over 200,000 steps, with a warm-up phase during the first 1,000 steps.
We use two 80-GB NVIDIA A100 GPUs configured for multi-GPU training using DistributedDataParallel (DDP) as implemented by the framework PyTorch (version 2.5.1, CUDA 12.4), with a batch size of 8. We do not divide the data for TriadMR-131K but use all the data to pretrain Triad and then save the model at 200,000 steps to serve as the initial parameters for downstream tasks.
For fine-tuning and validation of downstream tasks, we use the repository provided by VoCo v2~\cite{wu2024voco,wu2024large} (\url{https://github.com/Luffy03/Large-Scale-Medical}) and TransMorph~\cite{chen2022transmorph} (\url{https://github.com/junyuchen245/TransMorph_Transformer_for_Medical_Image_Registration}), respectively.

\subsection{Evaluation metrics}

We used several evaluation metrics to thoroughly assess the capabilities of our Triad model across different tasks. Accuracy is a primary metric used for evaluating the performance in medical-image classification, it is defined as the ratio of the number of correctly predicted samples to the total number of samples:

\begin{equation}
	\label{e:acc}
	\text{Accuracy} = \frac{TP + TN}{TP + TN + FP + FN}
\end{equation}
where TP (True Positives) denotes the number of samples correctly predicted as positive by the model; TN (True Negatives) denotes the number of samples correctly predicted as negative by the model; FP (False Positives) denotes the number of negative samples incorrectly predicted as positive by the model; FN (False Negatives) denotes the number of positive samples incorrectly predicted as negative by the model.

Dice Similarity Coefficient (DSC) is used to measure the overlap between two sets, which is widely used in medical image segmentation tasks:

\begin{equation}
	\label{e:dsc}
	\text{DSC} = \frac{2|X \cap Y|}{|X|+|Y|}
\end{equation}
where $X$ is the pixel set of the predicted segmentation result; $Y$ is the pixel set of the ground truth segmentation; $|X \cap Y|$ denotes the number of pixels contained in the intersection of $X$ and $Y$; $|X|$ and $|Y|$ denote the number of pixels of $X$ and $Y$ respectively. Equivalently, the DSC can be calculated based on the pixel category, which is expressed in pixel-by-pixel binary form by the predicted label $P$ and the ground truth label $G$:

\begin{equation}
	\label{e:dsc2}
	\text{DSC} = \frac{2\sum_{i}P_iG_i}{\sum_{i}P_i+\sum_{i}G_i}
\end{equation}
where $P_i$ is the predicted label for the $i$-th pixel value; $G_i$ is the $i$-th pixel value of the ground truth label. The value of DSC ranges between 0 and 1, where 1 indicates perfect overlap and 0 indicates no overlap.

\section{Data Availability}
All data in this study are publicly available and can be accessed from: 
fastMRI-Brain~\cite{knoll2020fastmri,zbontar2018fastmri} (\url{https://fastmri.med.nyu.edu/}),
UCSF-PDGM~\cite{calabrese2022university} (\url{https://www.cancerimagingarchive.net/collection/ucsf-pdgm/}),
ACRIN-DSC-MR-Brain~\cite{kinahan2019data} (\url{https://www.cancerimagingarchive.net/collection/acrin-dsc-mr-brain/}),
TCGA-GBM~\cite{scarpace2016cancer} (\url{https://www.cancerimagingarchive.net/collection/tcga-gbm/}),
UPENN-GBM~\cite{bakas2021multi} (\url{https://www.cancerimagingarchive.net/collection/upenn-gbm/}),
IvyGAP~\cite{pati2020multi} (\url{https://www.cancerimagingarchive.net/analysis-result/ivygap-radiomics/}),
ACRIN-FMISO-Brain~\cite{kinahan2018data} (\url{https://www.cancerimagingarchive.net/collection/acrin-fmiso-brain/}),
IXI (\url{https://brain-development.org/ixi-dataset/}),
CPTAC-GBM~\cite{CPTACGBM2018} (\url{https://www.cancerimagingarchive.net/collection/cptac-gbm/}),
ReMIND~\cite{juvekar2023brain} (\url{https://www.cancerimagingarchive.net/collection/remind/}),
REMBRANDT~\cite{scarpace2019data} (\url{https://www.cancerimagingarchive.net/collection/rembrandt/}),
VESTIBULAR-sHWANNOMA~\cite{Kujawa2023_Vestibular_Schwannoma} (\url{https://www.cancerimagingarchive.net/collection/vestibular-schwannoma-mc-rc/}),
GLIS-RT~\cite{shusharina2021glioma} (\url{https://www.cancerimagingarchive.net/collection/glis-rt/}),
Meningioma-SEG-CLASS~\cite{vassantachart2023segmentation} (\url{https://www.cancerimagingarchive.net/collection/meningioma-seg-class/}),
ISPY2~\cite{newitt2021acrin,Li2022_ISPY2} (\url{https://www.cancerimagingarchive.net/collection/ispy2/}),
ACRIN-6698~\cite{newitt2021acrin} (\url{https://www.cancerimagingarchive.net/collection/acrin-6698/}),
ACRIN-Contralateral-Breast-MR~\cite{Kinahan2021} (\url{https://www.cancerimagingarchive.net/collection/acrin-contralateral-breast-mr/}),
EA1141~\cite{comstock2023abbreviated} (\url{https://www.cancerimagingarchive.net/collection/ea1141/}),
Advanced-MRI-Breast-Lesions~\cite{Daniels2024} (\url{https://www.cancerimagingarchive.net/collection/advanced-mri-breast-lesions/}),
ISPY1~\cite{newitt2016multi} (\url{https://www.cancerimagingarchive.net/collection/ispy1/}),
TCGA-BRCA~\cite{lingle2016cancer} (\url{https://www.cancerimagingarchive.net/collection/tcga-brca/}),
Breast-MRI-NACT-Pilot~\cite{newitt2016single} (\url{https://www.cancerimagingarchive.net/collection/breast-mri-nact-pilot/}),
QIN Breast DCE-MRI~\cite{huang2014variations} (\url{https://www.cancerimagingarchive.net/collection/qin-breast-dce-mri/}),
fastMRI-Breast~\cite{knoll2020fastmri,zbontar2018fastmri} (\url{https://fastmri.med.nyu.edu/}),
BREAST-DIAGNOSIS~\cite{bloch2015breast} (\url{https://www.cancerimagingarchive.net/collection/breast-diagnosis/}),
PROSTATEx~\cite{litjens2017prostatex} (\url{https://www.cancerimagingarchive.net/collection/prostatex/}),
BIMCV-Prostate~\cite{Alzate-Grisales2024} (\url{https://bimcv.cipf.es/bimcv-projects/prostate/}),
PI-CAI Challenge~\cite{saha2024artificial} (\url{https://pi-cai.grand-challenge.org/}),
Prostate-MRI-US-Biopsy~\cite{natarajan2020prostate} (\url{https://www.cancerimagingarchive.net/collection/prostate-mri-us-biopsy/}),
CPTAC-UCEC~\cite{CPTAC2019} (\url{https://www.cancerimagingarchive.net/collection/cptac-ucec/}),
fastMRI-Prostate~\cite{knoll2020fastmri,zbontar2018fastmri} (\url{https://fastmri.med.nyu.edu/}),
PROSTATE-DIAGNOSIS~\cite{bloch2015data} (\url{https://www.cancerimagingarchive.net/collection/prostate-diagnosis/}),
Prostate Fused-MRI-Pathology~\cite{madabhushi2016fused} (\url{https://www.cancerimagingarchive.net/collection/prostate-fused-mri-pathology/}),
PROSTATE-MRI~\cite{choyke2016data} (\url{https://www.cancerimagingarchive.net/collection/prostate-mri/}),
TCGA-PRAD~\cite{Zuley2016} (\url{https://www.cancerimagingarchive.net/collection/tcga-prad/}),
Prostate-3T~\cite{litjens2015data} (\url{https://www.cancerimagingarchive.net/collection/prostate-3t/}),
MSD Challenge~\cite{antonelli2022medical} (\url{https://decathlon-10.grand-challenge.org/}),
BraTs21~\cite{baid2021rsna} (\url{http://braintumorsegmentation.org/}),
BreastDM~\cite{zhao2023breastdm} (\url{https://github.com/smallboy-code/Breast-cancer-dataset}),
Prostate158~\cite{adams2022prostate158} (\url{https://zenodo.org/records/6481141}),
MM-WHS-MRI~\cite{zhuang2018multivariate} (\url{https://zmiclab.github.io/zxh/0/mmwhs/}),
ATLAS-MRI~\cite{quinton2023tumour} (\url{https://atlas-challenge.u-bourgogne.fr/}),
Abdoman 1K~\cite{ma2021abdomenct} (\url{https://github.com/JunMa11/AbdomenCT-1K?tab=readme-ov-file}),
Kipa22~\cite{he2021meta} (\url{https://kipa22.grand-challenge.org/}),
OASIS~\cite{marcus2007open} (\url{https://sites.wustl.edu/oasisbrains/}),
ACDC~\cite{bernard2018deep} (\url{https://www.creatis.insa-lyon.fr/Challenge/acdc/databases.html}),
OrganMNIST3D~\cite{yang2023medmnist} (\url{https://github.com/MedMNIST/MedMNIST/tree/main}),
LUNA16~\cite{setio2017validation} (\url{https://luna16.grand-challenge.org/Data/}),
LLD-MMRI~\cite{lou2024sdrformer} (\url{https://github.com/LMMMEng/LLD-MMRI-Dataset}),
ANDI~\cite{jack2008alzheimer} (\url{https://adni.loni.usc.edu/data-samples/adni-data/neuroimaging/mri/mri-image-data-sets/}).
\section{Code Availability}
The pretrained and fine-tuned models, as well as source code for training, inference and data preprocessing, can be accessed at \url{https://github.com/wangshansong1/Triad}.
\newpage
\section*{Supplementary Materials: Tables and Figures}
\begin{table}[!h]
	\centering
	\caption{\textbf{Datasets used in Triad for pre-training with details.}}
	\label{table:t1} 
	\resizebox{\textwidth}{!}{
	\begin{tabular}{ccccc m{9cm}}
		\hline
		\textbf{Organ} & \textbf{Dataset} & \textbf{Series Number} & \textbf{Subjects} & \textbf{Studies} & \textbf{Availability} \\ \hline 
		\multirow{14}[28]{*}{Brain} & fastMRI-Brain & 23,082 & 8,165 & 23,082 & \url{https://fastmri.med.nyu.edu/} \\ \cline{2-6}
		~ & UCSF-PDGM & 6,998 & 501 & 6,998 & \url{https://www.cancerimagingarchive.net/collection/upenn-gbm/} \\\cline{2-6}
		~ & ACRIN-DSC-MR-Brain & 5,093 & 123 & 547 & \url{https://www.cancerimagingarchive.net/collection/acrin-dsc-mr-brain/} \\ \cline{2-6}
		~ & TCGA-GBM & 3,547 & 255 & 520 & \url{https://www.cancerimagingarchive.net/collection/tcga-gbm/} \\ \cline{2-6}
		~ & UPENN-GBM & 3,129 & 627 & 3,109 & \url{https://www.cancerimagingarchive.net/collection/upenn-gbm/} \\ \cline{2-6}
		~ & IvyGAP & 2,769 & 39 & 385 & \url{https://www.cancerimagingarchive.net/analysis-result/ivygap-radiomics/} \\ \cline{2-6}
		~ & ACRIN-FMISO-Brain & 2,071 & 42 & 187 & \url{https://www.cancerimagingarchive.net/collection/acrin-fmiso-brain/} \\ \cline{2-6}
		~ & IXI & 1,090 & 581 & 1,090 & \url{https://brain-development.org/ixi-dataset/} \\ \cline{2-6}
		~ & CPTAC-GBM & 958 & 63 & 127 & \url{https://www.cancerimagingarchive.net/collection/cptac-gbm/} \\\cline{2-6}
		~ & ReMIND & 638 & 114 & 227 & \url{https://www.cancerimagingarchive.net/collection/remind/} \\ \cline{2-6}
		~ & REMBRANDT & 499 & 90 & 97 & \url{https://www.cancerimagingarchive.net/collection/rembrandt/} \\ \cline{2-6}
		~ & VESTIBULAR-sHWANNOMA & 484 & 242 & 484 & \url{https://www.cancerimagingarchive.net/collection/vestibular-schwannoma-mc-rc/} \\ \cline{2-6}
		~ & GLIS-RT & 431 & 230 & 431 & \url{https://www.cancerimagingarchive.net/collection/glis-rt/} \\  \cline{2-6}
		~ & Meningioma-SEG-CLASS & 323 & 89 & 152 & \url{https://www.cancerimagingarchive.net/collection/meningioma-seg-class/} \\ \cline{2-6}
		 \hline
		\multirow{11}[22]{*}{Breast} & ISPY2 & 18,494 & 714 & 2,648 & \url{https://www.cancerimagingarchive.net/collection/ispy2/} \\ \cline{2-6}
		~ & ACRIN-6698 & 8,668 & 385 & 1,148 & \url{https://www.cancerimagingarchive.net/collection/acrin-6698/} \\ \cline{2-6}
		~ & ACRIN-Contralateral-Breast-MR & 5,906 & 788 & 875 & \url{https://www.cancerimagingarchive.net/collection/acrin-contralateral-breast-mr/} \\ \cline{2-6}
		~ & EA1141 & 3,526 & 500 & 953 & \url{https://www.cancerimagingarchive.net/collection/ea1141/} \\\cline{2-6}
		~ & Advanced-MRI-Breast-Lesions & 3,435 & 632 & 632 & \url{https://www.cancerimagingarchive.net/collection/advanced-mri-breast-lesions/} \\ \cline{2-6}
		~ & ISPY1 & 3,292 & 220 & 834 & \url{https://www.cancerimagingarchive.net/collection/ispy1/} \\ \cline{2-6}
		~ & TCGA-BRCA & 867 & 137 & 156 & \url{https://www.cancerimagingarchive.net/collection/tcga-brca/} \\ \cline{2-6}
		~ & Breast-MRI-NACT-Pilot & 694 & 64 & 201 & \url{https://www.cancerimagingarchive.net/collection/breast-mri-nact-pilot/} \\ \cline{2-6}
		~ & QIN Breast DCE-MRI & 632 & 10 & 20 & \url{https://www.cancerimagingarchive.net/collection/qin-breast-dce-mri/} \\ \cline{2-6}
		~ & fastMRI-Breast & 600 & 300 & 600 & \url{https://fastmri.med.nyu.edu/} \\\cline{2-6}
		~ & BREAST-DIAGNOSIS & 290 & 84 & 113 & \url{https://www.cancerimagingarchive.net/collection/breast-diagnosis/} \\ \cline{2-6}
		 \hline
		\multirow{11}[22]{*}{Prostate} & PROSTATEx & 17,313 & 346 & 351 & \url{https://www.cancerimagingarchive.net/collection/prostatex/} \\ \cline{2-6}
		~ & BIMCV-Prostate & 6,397 & 1,501 & 1,531 & \url{https://bimcv.cipf.es/bimcv-projects/prostate/} \\\cline{2-6}
		~ & PI-CAI Challenge & 5,995 & 1,476 & 5,995 & \url{https://pi-cai.grand-challenge.org/} \\ \cline{2-6}
		~ & Prostate-MRI-US-Biopsy & 1,994 & 837 & 1,184 & \url{https://www.cancerimagingarchive.net/collection/prostate-mri-us-biopsy/} \\ \cline{2-6}
		~ & CPTAC-UCEC & 778 & 36 & 38 & \url{https://www.cancerimagingarchive.net/collection/cptac-ucec/} \\ \cline{2-6}
		~ & fastMRI-Prostate & 625 & 312 & 625 & \url{https://fastmri.med.nyu.edu/} \\ \cline{2-6}
		~ & PROSTATE-DIAGNOSIS & 261 & 89 & 89 & \url{https://www.cancerimagingarchive.net/collection/prostate-diagnosis/} \\ \cline{2-6}
		~ & Prostate Fused-MRI-Pathology & 233 & 28 & 28 & \url{https://www.cancerimagingarchive.net/collection/prostate-fused-mri-pathology/} \\ \cline{2-6}
		~ & PROSTATE-MRI & 158 & 26 & 26 & \url{https://www.cancerimagingarchive.net/collection/prostate-mri/} \\ \cline{2-6}
		~ & TCGA-PRAD & 124 & 10 & 10 & \url{https://www.cancerimagingarchive.net/collection/tcga-prad/} \\ \cline{2-6}
		~ & Prostate-3T & 64 & 64 & 64 & \url{https://www.cancerimagingarchive.net/collection/prostate-3t/} \\ 	\cline{2-6}		
	 \hline
	 \multirow{1}[0]{*}{\textbf{Total}} & \textbf{TriadMR-131K} & \textbf{131,170} & \textbf{19,721} & \textbf{55,557} & ~ \\ 
	 \hline
	\end{tabular}}
\end{table}
\begin{table}[!t]
	\centering
	\caption{\textbf{The corresponding labels of the category numbers in Fig. \ref{fig:cls} b.}}
	\label{table:t2} 
	
	\begin{tabular}{ccl}
		\hline
		Dataset & No. & Label \\ \hline
		\multirow{2}[1]{*}{BreastDM} & 0 & Benign \\ 
		~ & 1 & Malignant \\ 
		\hline
		\multirow{2}[1]{*}{ADNI} & 0 & Cognitively Normal (CN) \\ 
		~ & 1 & Mild Cognitive Impairment (MCI) \\ 
		~ & 2 & Alzheimer’s Disease (AD) \\ 
		\hline
		\multirow{7}[1]{*}{LLD-MMRI} & 0 & Hepatocellular Carcinoma (HCC) \\ 
		~ & 1 & Intrahepatic Cholangio Carcinoma (ICC) \\ 
		~ & 2 & Hepatic Metastasis (HM) \\ 
		~ & 3 & Hepatic Cyst (HC) \\ 
		~ & 4 & Hepatic Hemangioma (HH) \\ 
		~ & 5 & Focal Nodular Hyperplasia (FNH) \\ 
		~ & 6 & Hepatic Abscess (HA) \\ 
		\hline
		\multirow{2}[1]{*}{LUNA16} & 0 & Non-nodule \\ 
		~ & 1 & Nodule \\ 
		\hline
		\multirow{11}[1]{*}{OrganMNIST3D} & 0 & Heart \\ 
		~ & 1 & Left Lung \\ 
		~ & 2 &  Right Lung \\ 
		~ & 3 & Liver \\ 
		~ & 4 & Spleen \\ 
		~ & 5 & Pancreas \\ 
		~ & 6 & Left Kidney \\ 
		~ & 7 & Right Kidney \\ 
		~ & 8 & Bladder \\ 
		~ & 9 & Left Femoral Head \\ 
		~ & 10 & Right Femoral Head \\ \hline
	\end{tabular}
\end{table}

\begin{table}[!ht]
	\centering
	\caption{\textbf{Parameter setting in the pre-training phase.}}
	\label{table:t3} 
	\begin{tabular}{cccc}
		\hline
		Encoder & Scale & Parameter & Value \\ \hline
		nnUNet & - & Learning rate & 0.0001 \\ 
		\hline
		\multirow{9}[1]{*}{Swin-Transformer} & \multirow{3}[1]{*}{Base} & Learning rate & 0.000001 \\ 
		~ & ~ & Feature size & 48 \\ 
		~ & ~ & Bottleneck Depth & 768 \\ 
		\cline{2-4}
		~ & \multirow{3}[1]{*}{Large} & Learning rate & 0.000001 \\ 
		~ & ~ & Feature size & 96 \\ 
		~ & ~ & Bottleneck Depth & 1,536 \\ 
		\cline{2-4}
		~ & \multirow{3}[1]{*}{Huge} & Learning rate & 0.000001 \\ 
		~ & ~ & Feature size & 192 \\ 
		~ & ~ & Bottleneck Depth & 3,072 \\ 
		\hline
		Common parameter & ~ & Value & ~ \\ 
		\hline
		Optimizer & & AdamW  & \\ 
		Number step & & 200,000 & \\ 
		Warmup step & & 1,000 & \\ 
		Learning rate schedule & & Cosine & \\ 
		Batch size & & 8 & \\ 
		Roi x,y,z & & 96 & \\ 
		\hline
	\end{tabular}
\end{table}
\clearpage

\begin{figure}[!t] 
	\centering
	\includegraphics[width=\textwidth]{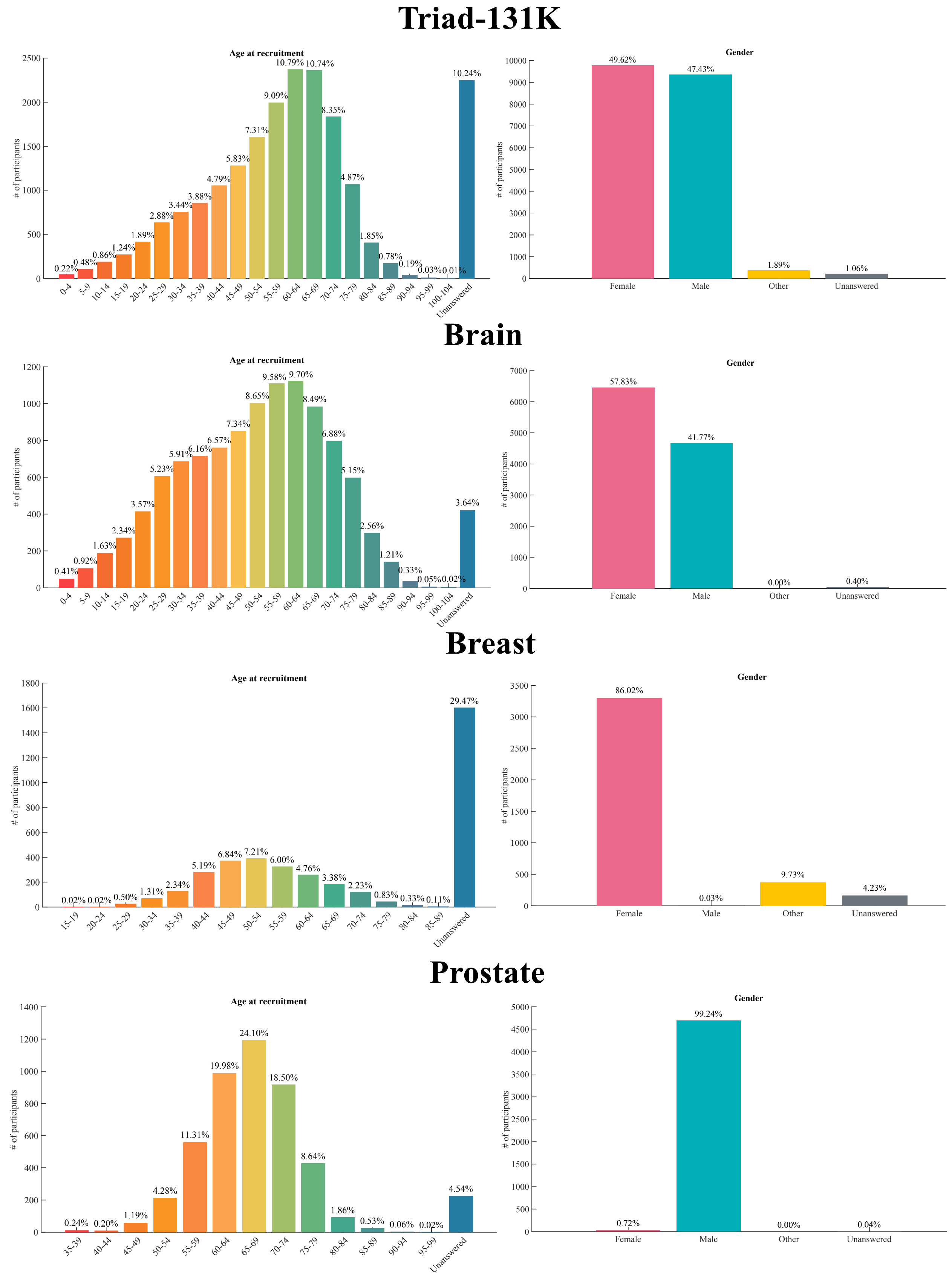}
	\caption{\textbf{Visual representation of the demographics of Triad-131K and its three subsets in this study.}}
	\label{fig:agegender} 
\end{figure}

\begin{figure}[!t] 
	\centering
	\includegraphics[width=\textwidth]{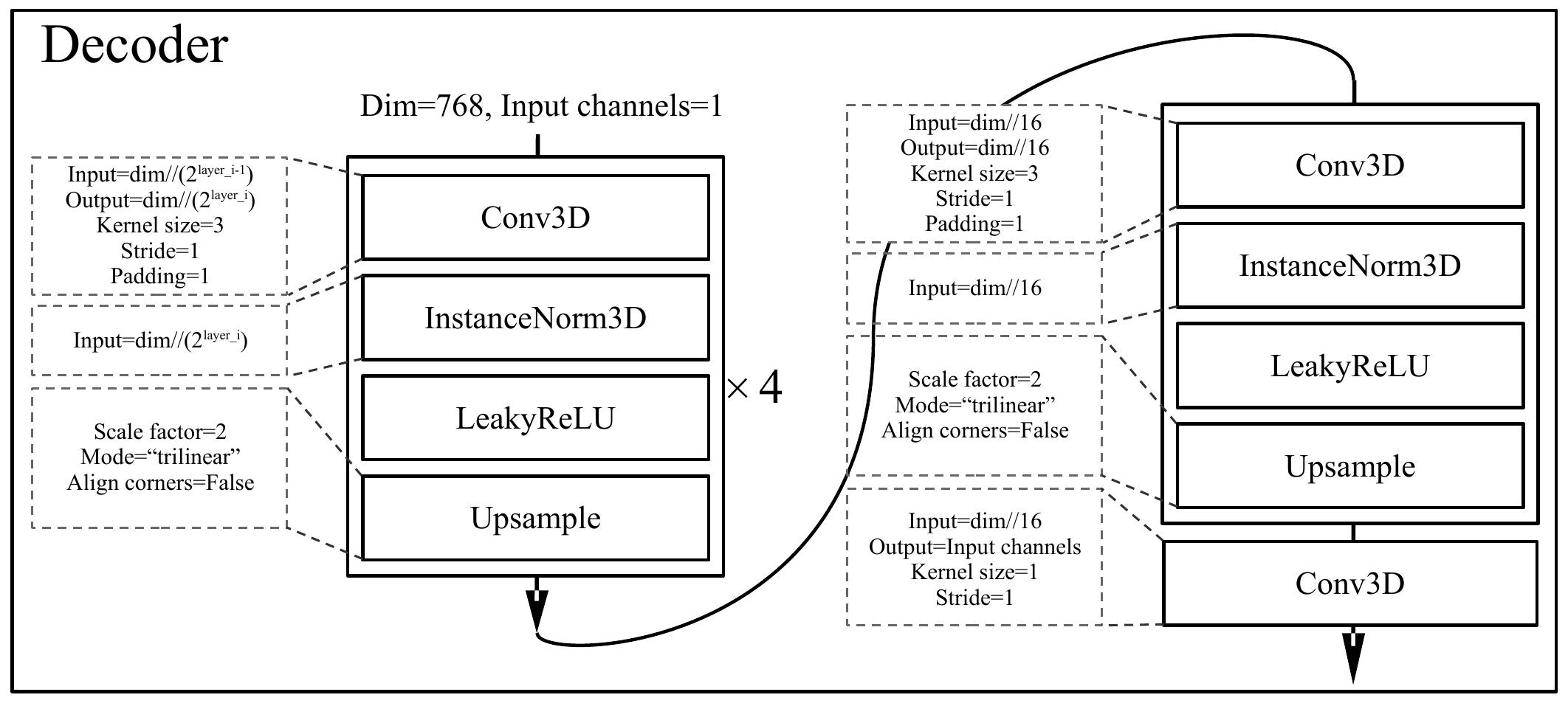}
	\caption{\textbf{Decoder architecture used uniformly in the pre-training phase.}}
	\label{fig:decoder} 
\end{figure}

\newpage

\subsection*{Acknowledgements} 

This research is supported in part by the National Institutes of Health under Award Number R56EB033332, R01EB032680, R01DE033512 and R01CA272991.

\subsection*{Contributions}

\textbf{Shansong Wang:} Writing-original draft, Methodology, Investigation,
Formal analysis, Data curation, Conceptualization.
\textbf{Mojtaba Safari:} Data curation, Writing-review \& editing. 
\textbf{Qiang Li:} Investigation, Writing-review \& editing. 
\textbf{Chih-Wei Chang:} Writing-review \& editing. 
\textbf{Richard LJ Qiu:} Writing-review \& editing. 
\textbf{Justin Roper:} Writing-review \& editing. 
\textbf{David S. Yu:} Writing-review \& editing. 
\textbf{Xiaofeng Yang:} Writing-review \& editing, Supervision, Resources, Project administration, Funding acquisition.

\bibliographystyle{unsrt}
\bibliography{newbib}

\begin{thebibliography}{10}

\bibitem{oecd_mri_units_2023}
Organisation for Economic Co-operation and Development.
\newblock Number of magnetic resonance imaging (mri) units in selected
  countries as of 2021, 2023.

\bibitem{shah2023review}
Aren Shah and Shima Aran.
\newblock A review of magnetic resonance (mr) safety: The essentials to patient
  safety.
\newblock {\em Cureus}, 15(10), 2023.

\bibitem{mri_number_worldwide}
Collective Minds.
\newblock Number of mri scans per year worldwide: Overview of global mri
  utilization, 2024.

\bibitem{https://doi.org/10.1002/mp.16844}
Mojtaba Safari, Xiaofeng Yang, Ali Fatemi, and Louis Archambault.
\newblock Mri motion artifact reduction using a conditional diffusion
  probabilistic model (mar-cdpm).
\newblock {\em Medical Physics}, 51(4):2598--2610, 2024.

\bibitem{Caron_2021_ICCV}
Mathilde Caron, Hugo Touvron, Ishan Misra, Herv\'e J\'egou, Julien Mairal,
  Piotr Bojanowski, and Armand Joulin.
\newblock Emerging properties in self-supervised vision transformers.
\newblock In {\em Proceedings of the IEEE/CVF International Conference on
  Computer Vision (ICCV)}, pages 9650--9660, October 2021.

\bibitem{he2022masked}
Kaiming He, Xinlei Chen, Saining Xie, Yanghao Li, Piotr Doll{\'a}r, and Ross
  Girshick.
\newblock Masked autoencoders are scalable vision learners.
\newblock In {\em Proceedings of the IEEE/CVF conference on computer vision and
  pattern recognition}, pages 16000--16009, 2022.

\bibitem{oquab2023dinov2}
Maxime Oquab, Timoth{\'e}e Darcet, Th{\'e}o Moutakanni, Huy Vo, Marc
  Szafraniec, Vasil Khalidov, Pierre Fernandez, Daniel Haziza, Francisco Massa,
  Alaaeldin El-Nouby, et~al.
\newblock Dinov2: Learning robust visual features without supervision.
\newblock {\em arXiv preprint arXiv:2304.07193}, 2023.

\bibitem{moor2023foundation}
Michael Moor, Oishi Banerjee, Zahra Shakeri~Hossein Abad, Harlan~M Krumholz,
  Jure Leskovec, Eric~J Topol, and Pranav Rajpurkar.
\newblock Foundation models for generalist medical artificial intelligence.
\newblock {\em Nature}, 616(7956):259--265, 2023.

\bibitem{chen2024towards}
Richard~J Chen, Tong Ding, Ming~Y Lu, Drew~FK Williamson, Guillaume Jaume,
  Andrew~H Song, Bowen Chen, Andrew Zhang, Daniel Shao, Muhammad Shaban, et~al.
\newblock Towards a general-purpose foundation model for computational
  pathology.
\newblock {\em Nature Medicine}, 30(3):850--862, 2024.

\bibitem{Zhao2024}
Theodore Zhao, Yu~Gu, Jianwei Yang, Naoto Usuyama, Ho~Hin Lee, Sid Kiblawi,
  Tristan Naumann, Jianfeng Gao, Angela Crabtree, Jacob Abel, Christine
  Moung-Wen, Brian Piening, Carlo Bifulco, Mu~Wei, Hoifung Poon, and Sheng
  Wang.
\newblock A foundation model for joint segmentation, detection and recognition
  of biomedical objects across nine modalities.
\newblock {\em Nature Methods}, November 2024.
\newblock Published online.

\bibitem{Safari_2024}
Mojtaba Safari, Xiaofeng Yang, Chih-Wei Chang, Richard L~J Qiu, Ali Fatemi, and
  Louis Archambault.
\newblock Unsupervised mri motion artifact disentanglement: introducing
  maudgan.
\newblock {\em Physics in Medicine \& Biology}, 69(11):115057, may 2024.

\bibitem{kim2019deep}
Sungyeon Kim, Minkyo Seo, Ivan Laptev, Minsu Cho, and Suha Kwak.
\newblock Deep metric learning beyond binary supervision.
\newblock In {\em Proceedings of the IEEE/CVF Conference on Computer Vision and
  Pattern Recognition}, pages 2288--2297, 2019.

\bibitem{perez2021torchio}
Fernando P{\'e}rez-Garc{\'\i}a, Rachel Sparks, and S{\'e}bastien Ourselin.
\newblock Torchio: a python library for efficient loading, preprocessing,
  augmentation and patch-based sampling of medical images in deep learning.
\newblock {\em Computer methods and programs in biomedicine}, 208:106236, 2021.

\bibitem{cardoso2022monai}
M~Jorge Cardoso, Wenqi Li, Richard Brown, Nic Ma, Eric Kerfoot, Yiheng Wang,
  Benjamin Murrey, Andriy Myronenko, Can Zhao, Dong Yang, et~al.
\newblock Monai: An open-source framework for deep learning in healthcare.
\newblock {\em arXiv preprint arXiv:2211.02701}, 2022.

\bibitem{https://doi.org/10.1002/mp.17675}
Mojtaba Safari, Zach Eidex, Shaoyan Pan, Richard L.~J. Qiu, and Xiaofeng Yang.
\newblock Self-supervised adversarial diffusion models for fast mri
  reconstruction.
\newblock {\em Medical Physics}, n/a(n/a).

\bibitem{wu2024voco}
Linshan Wu, Jiaxin Zhuang, and Hao Chen.
\newblock Voco: A simple-yet-effective volume contrastive learning framework
  for 3d medical image analysis.
\newblock In {\em Proceedings of the IEEE/CVF Conference on Computer Vision and
  Pattern Recognition}, pages 22873--22882, 2024.

\bibitem{li2024well}
Wenxuan Li, Alan Yuille, and Zongwei Zhou.
\newblock How well do supervised models transfer to 3d image segmentation.
\newblock In {\em The Twelfth International Conference on Learning
  Representations}, volume~1, 2024.

\bibitem{tang2022self}
Yucheng Tang, Dong Yang, Wenqi Li, Holger~R Roth, Bennett Landman, Daguang Xu,
  Vishwesh Nath, and Ali Hatamizadeh.
\newblock Self-supervised pre-training of swin transformers for 3d medical
  image analysis.
\newblock In {\em Proceedings of the IEEE/CVF conference on computer vision and
  pattern recognition}, pages 20730--20740, 2022.

\bibitem{ye2024continual}
Yiwen Ye, Yutong Xie, Jianpeng Zhang, Ziyang Chen, Qi~Wu, and Yong Xia.
\newblock Continual self-supervised learning: Towards universal multi-modal
  medical data representation learning.
\newblock In {\em Proceedings of the IEEE/CVF Conference on Computer Vision and
  Pattern Recognition}, pages 11114--11124, 2024.

\bibitem{gao2024training}
Yunhe Gao.
\newblock Training like a medical resident: Context-prior learning toward
  universal medical image segmentation.
\newblock In {\em Proceedings of the IEEE/CVF Conference on Computer Vision and
  Pattern Recognition}, pages 11194--11204, 2024.

\bibitem{isensee2021nnu}
Fabian Isensee, Paul~F Jaeger, Simon~AA Kohl, Jens Petersen, and Klaus~H
  Maier-Hein.
\newblock nnu-net: a self-configuring method for deep learning-based biomedical
  image segmentation.
\newblock {\em Nature methods}, 18(2):203--211, 2021.

\bibitem{cox2024brainsegfounder}
Joseph Cox, Peng Liu, Skylar~E Stolte, Yunchao Yang, Kang Liu, Kyle~B See,
  Huiwen Ju, and Ruogu Fang.
\newblock Brainsegfounder: towards 3d foundation models for neuroimage
  segmentation.
\newblock {\em Medical Image Analysis}, 97:103301, 2024.

\bibitem{kim2023empirical}
Heejong Kim, Victor~Ion Butoi, Adrian~V Dalca, and Mert~R Sabuncu.
\newblock Empirical analysis of a segmentation foundation model in prostate
  imaging.
\newblock In {\em International Conference on Medical Image Computing and
  Computer-Assisted Intervention}, pages 140--150. Springer, 2023.

\bibitem{blankemeier2024merlin}
Louis Blankemeier, Joseph~Paul Cohen, Ashwin Kumar, Dave Van~Veen, Syed
  Jamal~Safdar Gardezi, Magdalini Paschali, Zhihong Chen, Jean-Benoit
  Delbrouck, Eduardo Reis, Cesar Truyts, et~al.
\newblock Merlin: A vision language foundation model for 3d computed
  tomography.
\newblock {\em arXiv preprint arXiv:2406.06512}, 2024.

\bibitem{zhang2024generalist}
Kai Zhang, Rong Zhou, Eashan Adhikarla, Zhiling Yan, Yixin Liu, Jun Yu,
  Zhengliang Liu, Xun Chen, Brian~D Davison, Hui Ren, et~al.
\newblock A generalist vision--language foundation model for diverse biomedical
  tasks.
\newblock {\em Nature Medicine}, pages 1--13, 2024.

\bibitem{baid2021rsna}
Ujjwal Baid, Satyam Ghodasara, Suyash Mohan, Michel Bilello, Evan Calabrese,
  Errol Colak, Keyvan Farahani, Jayashree Kalpathy-Cramer, Felipe~C Kitamura,
  Sarthak Pati, et~al.
\newblock The rsna-asnr-miccai brats 2021 benchmark on brain tumor segmentation
  and radiogenomic classification.
\newblock {\em arXiv preprint arXiv:2107.02314}, 2021.

\bibitem{antonelli2022medical}
Michela Antonelli, Annika Reinke, Spyridon Bakas, Keyvan Farahani, Annette
  Kopp-Schneider, Bennett~A Landman, Geert Litjens, Bjoern Menze, Olaf
  Ronneberger, Ronald~M Summers, et~al.
\newblock The medical segmentation decathlon.
\newblock {\em Nature communications}, 13(1):4128, 2022.

\bibitem{zhao2023breastdm}
Xiaoming Zhao, Yuehui Liao, Jiahao Xie, Xiaxia He, Shiqing Zhang, Guoyu Wang,
  Jiangxiong Fang, Hongsheng Lu, and Jun Yu.
\newblock Breastdm: A dce-mri dataset for breast tumor image segmentation and
  classification.
\newblock {\em Computers in Biology and Medicine}, 164:107255, 2023.

\bibitem{adams2022prostate158}
Lisa~C Adams, Marcus~R Makowski, G{\"u}nther Engel, Maximilian Rattunde, Felix
  Busch, Patrick Asbach, Stefan~M Niehues, Shankeeth Vinayahalingam, Bram van
  Ginneken, Geert Litjens, et~al.
\newblock Prostate158-an expert-annotated 3t mri dataset and algorithm for
  prostate cancer detection.
\newblock {\em Computers in Biology and Medicine}, 148:105817, 2022.

\bibitem{zhuang2018multivariate}
Xiahai Zhuang.
\newblock Multivariate mixture model for myocardial segmentation combining
  multi-source images.
\newblock {\em IEEE transactions on pattern analysis and machine intelligence},
  41(12):2933--2946, 2018.

\bibitem{quinton2023tumour}
F{\'e}lix Quinton, Romain Popoff, Beno{\^\i}t Presles, Sarah Leclerc, Fabrice
  Meriaudeau, Guillaume Nodari, Olivier Lopez, Julie Pellegrinelli, Olivier
  Chevallier, Dominique Ginhac, et~al.
\newblock A tumour and liver automatic segmentation (atlas) dataset on
  contrast-enhanced magnetic resonance imaging for hepatocellular carcinoma.
\newblock {\em Data}, 8(5):79, 2023.

\bibitem{ma2021abdomenct}
Jun Ma, Yao Zhang, Song Gu, Cheng Zhu, Cheng Ge, Yichi Zhang, Xingle An,
  Congcong Wang, Qiyuan Wang, Xin Liu, et~al.
\newblock Abdomenct-1k: Is abdominal organ segmentation a solved problem?
\newblock {\em IEEE Transactions on Pattern Analysis and Machine Intelligence},
  44(10):6695--6714, 2021.

\bibitem{he2021meta}
Yuting He, Guanyu Yang, Jian Yang, Rongjun Ge, Youyong Kong, Xiaomei Zhu,
  Shaobo Zhang, Pengfei Shao, Huazhong Shu, Jean-Louis Dillenseger, et~al.
\newblock Meta grayscale adaptive network for 3d integrated renal structures
  segmentation.
\newblock {\em Medical image analysis}, 71:102055, 2021.

\bibitem{ghesu2022contrastive}
Florin~C Ghesu, Bogdan Georgescu, Awais Mansoor, Youngjin Yoo, Dominik Neumann,
  Pragneshkumar Patel, Reddappagari~Suryanarayana Vishwanath, James~M Balter,
  Yue Cao, Sasa Grbic, et~al.
\newblock Contrastive self-supervised learning from 100 million medical images
  with optional supervision.
\newblock {\em Journal of Medical Imaging}, 9(6):064503--064503, 2022.

\bibitem{amadou2024echoapex}
Abdoul~Aziz Amadou, Yue Zhang, Sebastien Piat, Paul Klein, Ingo Schmuecking,
  Tiziano Passerini, and Puneet Sharma.
\newblock Echoapex: A general-purpose vision foundation model for
  echocardiography.
\newblock {\em arXiv preprint arXiv:2410.11092}, 2024.

\bibitem{shi2025we}
Baifeng Shi, Ziyang Wu, Maolin Mao, Xin Wang, and Trevor Darrell.
\newblock When do we not need larger vision models?
\newblock In {\em European Conference on Computer Vision}, pages 444--462.
  Springer, 2025.

\bibitem{mei2024bigger}
Kangfu Mei, Zhengzhong Tu, Mauricio Delbracio, Hossein Talebi, Vishal~M Patel,
  and Peyman Milanfar.
\newblock Bigger is not always better: Scaling properties of latent diffusion
  models.
\newblock {\em arXiv preprint arXiv:2404.01367}, 3, 2024.

\bibitem{he2023swinunetr}
Yufan He, Vishwesh Nath, Dong Yang, Yucheng Tang, Andriy Myronenko, and Daguang
  Xu.
\newblock Swinunetr-v2: Stronger swin transformers with stagewise convolutions
  for 3d medical image segmentation.
\newblock In {\em International Conference on Medical Image Computing and
  Computer-Assisted Intervention}, pages 416--426. Springer, 2023.

\bibitem{dosovitskiy2020image}
Alexey Dosovitskiy.
\newblock An image is worth 16x16 words: Transformers for image recognition at
  scale.
\newblock {\em arXiv preprint arXiv:2010.11929}, 2020.

\bibitem{jack2008alzheimer}
Clifford~R Jack~Jr, Matt~A Bernstein, Nick~C Fox, Paul Thompson, Gene
  Alexander, Danielle Harvey, Bret Borowski, Paula~J Britson, Jennifer
  L.~Whitwell, Chadwick Ward, et~al.
\newblock The alzheimer's disease neuroimaging initiative (adni): Mri methods.
\newblock {\em Journal of Magnetic Resonance Imaging: An Official Journal of
  the International Society for Magnetic Resonance in Medicine},
  27(4):685--691, 2008.

\bibitem{yang2023medmnist}
Jiancheng Yang, Rui Shi, Donglai Wei, Zequan Liu, Lin Zhao, Bilian Ke,
  Hanspeter Pfister, and Bingbing Ni.
\newblock Medmnist v2-a large-scale lightweight benchmark for 2d and 3d
  biomedical image classification.
\newblock {\em Scientific Data}, 10(1):41, 2023.

\bibitem{setio2017validation}
Arnaud Arindra~Adiyoso Setio, Alberto Traverso, Thomas De~Bel, Moira~SN Berens,
  Cas Van Den~Bogaard, Piergiorgio Cerello, Hao Chen, Qi~Dou, Maria~Evelina
  Fantacci, Bram Geurts, et~al.
\newblock Validation, comparison, and combination of algorithms for automatic
  detection of pulmonary nodules in computed tomography images: the luna16
  challenge.
\newblock {\em Medical image analysis}, 42:1--13, 2017.

\bibitem{lou2024sdrformer}
Meng Lou, Hanning Ying, Xiaoqing Liu, Hong-Yu Zhou, Yuqing Zhang, and Yizhou
  Yu.
\newblock Sdr-former: A siamese dual-resolution transformer for liver lesion
  classification using 3d multi-phase imaging.
\newblock {\em arXiv preprint arXiv:2402.17246}, 2024.

\bibitem{cao2022swin}
Hu~Cao, Yueyue Wang, Joy Chen, Dongsheng Jiang, Xiaopeng Zhang, Qi~Tian, and
  Manning Wang.
\newblock Swin-unet: Unet-like pure transformer for medical image segmentation.
\newblock In {\em European conference on computer vision}, pages 205--218.
  Springer, 2022.

\bibitem{heydarian2022mlcm}
Mohammadreza Heydarian, Thomas~E Doyle, and Reza Samavi.
\newblock Mlcm: Multi-label confusion matrix.
\newblock {\em Ieee Access}, 10:19083--19095, 2022.

\bibitem{chen2022transmorph}
Junyu Chen, Eric~C Frey, Yufan He, William~P Segars, Ye~Li, and Yong Du.
\newblock Transmorph: Transformer for unsupervised medical image registration.
\newblock {\em Medical image analysis}, 82:102615, 2022.

\bibitem{hatamizadeh2021swin}
Ali Hatamizadeh, Vishwesh Nath, Yucheng Tang, Dong Yang, Holger~R Roth, and
  Daguang Xu.
\newblock Swin unetr: Swin transformers for semantic segmentation of brain
  tumors in mri images.
\newblock In {\em International MICCAI brainlesion workshop}, pages 272--284.
  Springer, 2021.

\bibitem{Brain-Development_IXI_2019}
{Brain-Development}.
\newblock Brain-development ixi dataset, 2019.
\newblock Accessed: 2025-02-04.

\bibitem{krentzel2023clem}
Daniel Krentzel, Matou{\v{s}} Elphick, Marie-Charlotte Domart, Christopher~J
  Peddie, Romain~F Laine, Cameron Shand, Ricardo Henriques, Lucy~M Collinson,
  and Martin~L Jones.
\newblock Clem-reg: An automated point cloud based registration algorithm for
  correlative light and volume electron microscopy.
\newblock {\em BioRxiv}, pages 2023--05, 2023.

\bibitem{bernard2018deep}
Olivier Bernard, Alain Lalande, Clement Zotti, Frederick Cervenansky, Xin Yang,
  Pheng-Ann Heng, Irem Cetin, Karim Lekadir, Oscar Camara, Miguel
  Angel~Gonzalez Ballester, et~al.
\newblock Deep learning techniques for automatic mri cardiac multi-structures
  segmentation and diagnosis: is the problem solved?
\newblock {\em IEEE transactions on medical imaging}, 37(11):2514--2525, 2018.

\bibitem{zhuang2024mim}
Jiaxin Zhuang, Linshan Wu, Qiong Wang, Varut Vardhanabhuti, Lin Luo, and Hao
  Chen.
\newblock Mim: Mask in mask self-supervised pre-training for 3d medical image
  analysis.
\newblock {\em arXiv preprint arXiv:2404.15580}, 2024.

\bibitem{jiang2023anatomical}
Yankai Jiang, Mingze Sun, Heng Guo, Xiaoyu Bai, Ke~Yan, Le~Lu, and Minfeng Xu.
\newblock Anatomical invariance modeling and semantic alignment for
  self-supervised learning in 3d medical image analysis.
\newblock In {\em Proceedings of the IEEE/CVF International Conference on
  Computer Vision}, pages 15859--15869, 2023.

\bibitem{yang2024freemask}
Lihe Yang, Xiaogang Xu, Bingyi Kang, Yinghuan Shi, and Hengshuang Zhao.
\newblock Freemask: Synthetic images with dense annotations make stronger
  segmentation models.
\newblock {\em Advances in Neural Information Processing Systems}, 36, 2024.

\bibitem{kirillov2023segment}
Alexander Kirillov, Eric Mintun, Nikhila Ravi, Hanzi Mao, Chloe Rolland, Laura
  Gustafson, Tete Xiao, Spencer Whitehead, Alexander~C Berg, Wan-Yen Lo, et~al.
\newblock Segment anything.
\newblock In {\em Proceedings of the IEEE/CVF International Conference on
  Computer Vision}, pages 4015--4026, 2023.

\bibitem{ni2021large}
Jianmo Ni, Chen Qu, Jing Lu, Zhuyun Dai, Gustavo~Hern{\'a}ndez {\'A}brego,
  Ji~Ma, Vincent~Y Zhao, Yi~Luan, Keith~B Hall, Ming-Wei Chang, et~al.
\newblock Large dual encoders are generalizable retrievers.
\newblock {\em arXiv preprint arXiv:2112.07899}, 2021.

\bibitem{radford2021learning}
Alec Radford, Jong~Wook Kim, Chris Hallacy, Aditya Ramesh, Gabriel Goh,
  Sandhini Agarwal, Girish Sastry, Amanda Askell, Pamela Mishkin, Jack Clark,
  et~al.
\newblock Learning transferable visual models from natural language
  supervision.
\newblock In {\em International conference on machine learning}, pages
  8748--8763. PMLR, 2021.

\bibitem{majee2024enhancing}
Arindam Majee, Avisek Gupta, Sourav Raha, and Swagatam Das.
\newblock Enhancing mri-based classification of alzheimer's disease with
  explainable 3d hybrid compact convolutional transformers.
\newblock {\em arXiv preprint arXiv:2403.16175}, 2024.

\bibitem{he2023neural}
Xinzi He, Alan~Q Wang, and Mert~R Sabuncu.
\newblock Neural pre-processing: A learning framework for end-to-end brain mri
  pre-processing.
\newblock In {\em International Conference on Medical Image Computing and
  Computer-Assisted Intervention}, pages 258--267. Springer, 2023.

\bibitem{fischl2012freesurfer}
B~Fischl.
\newblock Freesurfer. neuroimage, 20 years of fmri 62, 774--781, 2012.

\bibitem{loshchilov2017decoupled}
I~Loshchilov.
\newblock Decoupled weight decay regularization.
\newblock {\em arXiv preprint arXiv:1711.05101}, 2017.

\bibitem{wu2024large}
Linshan Wu, Jiaxin Zhuang, and Hao Chen.
\newblock Large-scale 3d medical image pre-training with geometric context
  priors.
\newblock {\em arXiv preprint arXiv:2410.09890}, 2024.

\bibitem{knoll2020fastmri}
Florian Knoll, Jure Zbontar, Anuroop Sriram, Matthew~J Muckley, Mary Bruno,
  Aaron Defazio, Marc Parente, Krzysztof~J Geras, Joe Katsnelson, Hersh
  Chandarana, et~al.
\newblock fastmri: A publicly available raw k-space and dicom dataset of knee
  images for accelerated mr image reconstruction using machine learning.
\newblock {\em Radiology: Artificial Intelligence}, 2(1):e190007, 2020.

\bibitem{zbontar2018fastmri}
Jure Zbontar, Florian Knoll, Anuroop Sriram, Tullie Murrell, Zhengnan Huang,
  Matthew~J Muckley, Aaron Defazio, Ruben Stern, Patricia Johnson, Mary Bruno,
  et~al.
\newblock fastmri: An open dataset and benchmarks for accelerated mri.
\newblock {\em arXiv preprint arXiv:1811.08839}, 2018.

\bibitem{calabrese2022university}
E~Calabrese, J~Villanueva-Meyer, J~Rudie, A~Rauschecker, U~Baid, S~Bakas,
  S~Cha, J~Mongan, and C~Hess.
\newblock The university of california san francisco preoperative diffuse
  glioma mri (ucsf-pdgm)(version 4)[dataset].
\newblock {\em The Cancer Imaging Archive. DOI}, 10, 2022.

\bibitem{kinahan2019data}
P~Kinahan, M~Muzi, B~Bialecki, B~Herman, and L~Coombs.
\newblock Data from acrin-dsc-mr-brain.
\newblock {\em The Cancer Imaging Archive}, 2019.

\bibitem{scarpace2016cancer}
Lisa Scarpace, Tom Mikkelsen, Soonmee Cha, Sujaya Rao, Sangeeta Tekchandani,
  David Gutman, Joel~H Saltz, Bradley~J Erickson, Nancy Pedano, Adam~E
  Flanders, et~al.
\newblock The cancer genome atlas glioblastoma multiforme collection
  (tcga-gbm).
\newblock {\em The Cancer Imaging Archive}, 2016.

\bibitem{bakas2021multi}
Spyridon Bakas, Chiharu Sako, Hamed Akbari, M~Bilello, A~Sotiras, G~Shukla,
  et~al.
\newblock Multi-parametric magnetic resonance imaging (mpmri) scans for de novo
  glioblastoma (gbm) patients from the university of pennsylvania health system
  (upenn-gbm).
\newblock {\em The Cancer Imaging Archive (TCIA) Public Access}, 2021.

\bibitem{pati2020multi}
S~Pati, R~Verma, H~Akbari, et~al.
\newblock Multi-institutional paired expert segmentations and radiomic features
  of the ivy gap dataset.
\newblock {\em The Cancer Imaging Archive}, 10, 2020.

\bibitem{kinahan2018data}
P~Kinahan, M~Muzi, B~Bialecki, and L.~Coombs.
\newblock Data from acrin-fmiso-brain.
\newblock {\em The Cancer Imaging Archive}, 2018.

\bibitem{CPTACGBM2018}
National Cancer Institute Clinical Proteomic Tumor Analysis~Consortium (CPTAC).
\newblock The clinical proteomic tumor analysis consortium glioblastoma
  multiforme collection (cptac-gbm).
\newblock {\em The Cancer Imaging Archive}, 2018.

\bibitem{juvekar2023brain}
P~Juvekar et~al.
\newblock The brain resection multimodal imaging database (remind).
\newblock {\em The Cancer Imaging Archive}, 2023.

\bibitem{scarpace2019data}
L~Scarpace, AE~Flanders, R~Jain, T~Mikkelsen, and DW~Andrews.
\newblock Data from rembrandt [data set]. the cancer imaging archive, 2019.

\bibitem{Kujawa2023_Vestibular_Schwannoma}
A.~Kujawa, R.~Dorent, N.~Wijethilake, S.~Connor, S.~Thomson, M.~Ivory,
  R.~Bradford, N.~Kitchen, S.~Bisdas, S.~Ourselin, T.~Vercauteren, and
  J.~Shapey.
\newblock Segmentation of vestibular schwannoma from magnetic resonance
  imaging: An annotated multi-center routine clinical dataset
  (vestibular-schwannoma-mc-rc).
\newblock {\em The Cancer Imaging Archive}, 2023.

\bibitem{shusharina2021glioma}
N~Shusharina and T~Bortfeld.
\newblock Glioma image segmentation for radiotherapy: Rt targets, barriers to
  cancer spread, and organs at risk [data set].
\newblock {\em The Cancer Imaging Archive}, 2021.

\bibitem{vassantachart2023segmentation}
A~Vassantachart, Y~Cao, Z~Shen, K~Cheng, M~Gribble, JC~Ye, G~Zada, K~Hurth,
  A~Mathew, S~Guzman, et~al.
\newblock Segmentation and classification of grade i and ii meningiomas from
  magnetic resonance imaging: an open annotated dataset (meningioma-seg-class).
\newblock {\em Cancer Imaging Arch}, 10, 2023.

\bibitem{newitt2021acrin}
David~C Newitt, SC~Partridge, Z~Zhang, J~Gibbs, T~Chenevert, M~Rosen, P~Bolan,
  H~Marques, J~Romanoff, L~Cimino, et~al.
\newblock Acrin 6698/i-spy2 breast dwi.
\newblock {\em The Cancer Imaging Archive}, 2021.

\bibitem{Li2022_ISPY2}
W.~Li, D.~C. Newitt, J.~Gibbs, L.~J. Wilmes, E.~F. Jones, V.~A. Arasu,
  F.~Strand, N.~Onishi, A.~A.-T. Nguyen, J.~Kornak, B.~N. Joe, E.~R. Price,
  H.~Ojeda-Fournier, M.~Eghtedari, K.~W. Zamora, S.~A. Woodard, H.~Umphrey,
  W.~Bernreuter, M.~Nelson, and N.~M. Hylton.
\newblock I-spy 2 breast dynamic contrast enhanced mri trial (ispy2).
\newblock {\em The Cancer Imaging Archive}, 2022.

\bibitem{Kinahan2021}
P.~Kinahan, M.~Muzi, B.~Bialecki, B.~Herman, and L.~Coombs.
\newblock {ACRIN-Contralateral-Breast-MR (ACRIN 6667) [Data set]}, 2021.

\bibitem{comstock2023abbreviated}
CE~Comstock, C~Gatsonis, GM~Newstead, BS~Snyder, IF~Gareen, JT~Bergin,
  H~Rahbar, JS~Sung, C~Jacobs, JA~Harvey, et~al.
\newblock Abbreviated breast mri and digital tomosynthesis mammography in
  screening women with dense breasts (ea1141).
\newblock {\em The Cancer Imaging Archive: Little Rock, AR, USA}, 2023.

\bibitem{Daniels2024}
D.~Daniels, D.~Last, K.~Cohen, Y.~Mardor, and M.~Sklair-Levy.
\newblock {Standard and Delayed Contrast-Enhanced MRI of Malignant and Benign
  Breast Lesions with Histological and Clinical Supporting Data
  (Advanced-MRI-Breast-Lesions) (Version 2) [Dataset]}, 2024.

\bibitem{newitt2016multi}
David Newitt, Nola Hylton, et~al.
\newblock Multi-center breast dce-mri data and segmentations from patients in
  the i-spy 1/acrin 6657 trials.
\newblock {\em Cancer Imaging Arch}, 10(7), 2016.

\bibitem{lingle2016cancer}
Wilma Lingle, Bradley~J Erickson, Margarita~L Zuley, Rose Jarosz, Ermelinda
  Bonaccio, Joe Filippini, Jose~M Net, Len Levi, Elizabeth~A Morris, Gloria~G
  Figler, et~al.
\newblock The cancer genome atlas breast invasive carcinoma collection
  (tcga-brca), 2016.

\bibitem{newitt2016single}
David Newitt and Nola Hylton.
\newblock Single site breast dce-mri data and segmentations from patients
  undergoing neoadjuvant chemotherapy.
\newblock {\em The Cancer Imaging Archive}, 2, 2016.

\bibitem{huang2014variations}
Wei Huang, Xin Li, Yiyi Chen, Xia Li, Ming-Ching Chang, Matthew~J Oborski,
  Dariya~I Malyarenko, Mark Muzi, Guido~H Jajamovich, Andriy Fedorov, et~al.
\newblock Variations of dynamic contrast-enhanced magnetic resonance imaging in
  evaluation of breast cancer therapy response: a multicenter data analysis
  challenge.
\newblock {\em Translational oncology}, 7(1):153--166, 2014.

\bibitem{bloch2015breast}
B~Nicolas Bloch, Ashali Jain, and CC~Jaffe.
\newblock Breast-diagnosis [dataset].
\newblock {\em The Cancer Imaging Archive}, 2015.

\bibitem{litjens2017prostatex}
Geert Litjens, Oscar Debats, Jelle Barentsz, Nico Karssemeijer, and Henkjan
  Huisman.
\newblock Prostatex challenge data.
\newblock {\em The cancer imaging archive}, 10:K9TCIA, 2017.

\bibitem{Alzate-Grisales2024}
J.~A. Alzate-Grisales and M.~de~la Iglesia~Vaya.
\newblock {BIMCV-Prostate-Dataset V1}, Aug 2024.

\bibitem{saha2024artificial}
Anindo Saha, Joeran~S Bosma, Jasper~J Twilt, Bram van Ginneken, Anders
  Bjartell, Anwar~R Padhani, David Bonekamp, Geert Villeirs, Georg Salomon,
  Gianluca Giannarini, et~al.
\newblock Artificial intelligence and radiologists in prostate cancer detection
  on mri (pi-cai): an international, paired, non-inferiority, confirmatory
  study.
\newblock {\em The Lancet Oncology}, 2024.

\bibitem{natarajan2020prostate}
S~Natarajan, A~Priester, D~Margolis, J~Huang, and L~Marks.
\newblock Prostate mri and ultrasound with pathology and coordinates of tracked
  biopsy (prostate-mri-us-biopsy).
\newblock {\em The Cancer Imaging Archive}, 10:7937, 2020.

\bibitem{CPTAC2019}
{National Cancer Institute Clinical Proteomic Tumor Analysis Consortium
  (CPTAC)}.
\newblock {The Clinical Proteomic Tumor Analysis Consortium Uterine Corpus
  Endometrial Carcinoma Collection (CPTAC-UCEC) (Version 12) [Data set]}, 2019.

\bibitem{bloch2015data}
B~Nicolas Bloch, Ashali Jain, and C~Carl Jaffe.
\newblock Data from prostate-diagnosis.
\newblock {\em the cancer imaging archive}, 9(10.7937), 2015.

\bibitem{madabhushi2016fused}
Anant Madabhushi and Michael Feldman.
\newblock Fused radiology-pathology prostate dataset.
\newblock {\em The Cancer Imaging Archive}, 9, 2016.

\bibitem{choyke2016data}
P~Choyke, B~Turkbey, P~Pinto, M~Merino, and B~Wood.
\newblock Data from prostate-mri.
\newblock {\em The Cancer Imaging Archive}, 9:6, 2016.

\bibitem{Zuley2016}
M.~L. Zuley, R.~Jarosz, B.~F. Drake, D.~Rancilio, A.~Klim, K.~Rieger-Christ,
  and J.~Lemmerman.
\newblock {The Cancer Genome Atlas Prostate Adenocarcinoma Collection
  (TCGA-PRAD) (Version 4) [Data set]}, 2016.

\bibitem{litjens2015data}
Geert Litjens, Jurgen Futterer, and Henkjan Huisman.
\newblock Data from prostate-3t, 2015.

\bibitem{marcus2007open}
Daniel~S Marcus, Tracy~H Wang, Jamie Parker, John~G Csernansky, John~C Morris,
  and Randy~L Buckner.
\newblock Open access series of imaging studies (oasis): cross-sectional mri
  data in young, middle aged, nondemented, and demented older adults.
\newblock {\em Journal of cognitive neuroscience}, 19(9):1498--1507, 2007.

\end{thebibliography}

\newpage
\appendix


\end{document}